%% file: neurips_2025.tex
\documentclass{article}

\input{preamble}

    \usepackage[final]{neurips_2025}

\usepackage[utf8]{inputenc} 
\usepackage[T1]{fontenc}    
\usepackage[pagebackref,breaklinks,colorlinks]{hyperref}
\usepackage{url}            
\usepackage{booktabs}       
\usepackage{amsfonts}       
\usepackage{nicefrac}       
\usepackage{microtype}      
\usepackage{xcolor}         
\newcommand{\Checkmark}{\ding{51}}  
\newcommand{\XSolidBrush}{\ding{55}}  
\usepackage{pifont}
\usepackage{graphicx}
\usepackage{tikz}
\usepackage{amsmath}
\usepackage{listings}
\usepackage{tcolorbox}

\usepackage[capitalise]{cleveref}

\title{\OurTitle}

\author{
    Zuwei Long$^{1, \dagger}$,
    Yunhang Shen$^{1, \dagger, \spadesuit}$,
    Chaoyou Fu$^{2,\star}$,
    Heting Gao$^{1}$,
    Lijiang Li$^{2}$ \\
    \and
    Peixian Chen$^{1}$,  
    Mengdan Zhang$^{1}$,
    Hang Shao$^{1}$,
    Jian Li$^{1}$,
    Jinlong Peng$^{1}$ \\
    \and
    Haoyu Cao$^{1}$,
    Ke Li$^{1}$, 
    Rongrong Ji$^{3}$,
    Xing Sun$^{1,\star}$
    \vspace{6pt}
    \\
    $^{1}$Tencent Youtu Lab, $^{2}$Nanjing University, $^{3}$Xiamen University
    \\
    \footnotesize{
    $^{\dagger}$~Equal Contribution \;
    $^{\spadesuit}$~Project Leader \;
    $^{\star}$~Corresponding Author \;
    }
    \\ \\
    \url{https://github.com/VITA-MLLM/VITA-Audio}
}

\begin{document}

\maketitle
\input{sec/0_abstract}

\input{sec/1_intro}

\input{sec/2_related}

\input{sec/3_method}
\input{sec/4_experiment}
\input{sec/5_conclusion}

\section*{Acknowledgments}
This work is partially funded by National Natural Science Foundation of China (Grant No. 62506158 and No. 62441234), and CCF-Tencent Rhino-Bird Open Research Fund.

\input{sec/append}

\clearpage
\newpage
{
\small
\bibliographystyle{plain}
\bibliography{library_format.bib}
}

\end{document}

%% file: preamble.tex
\usepackage{url}

\usepackage{subcaption}
\usepackage{caption}
\usepackage{newfloat}
\usepackage{listings}
\usepackage{xspace}

\usepackage{graphbox}
\usepackage{graphicx}
\usepackage{multirow}
\usepackage{wrapfig}
\usepackage{tikz}
\usepackage{float}

\usepackage{adjustbox}
\usepackage{booktabs}
\usepackage{stackengine}
\usepackage{tablefootnote}
\usepackage{threeparttable}

\usepackage{amsmath}
\usepackage{amssymb}
\usepackage{bbm}
\usepackage{dsfont}
\usepackage{mathtools}

\usepackage{color}
\usepackage{colortbl}
\usepackage[normalem]{ulem}
\usepackage{xcolor}

\usepackage{CJKutf8}

\newcommand{\hfilll}{\hspace{0pt plus 1filll}}

\makeatletter
\DeclareRobustCommand\onedot{\futurelet\@let@token\@onedot}
\def\@onedot{\ifx\@let@token.\else.\null\fi\xspace}

\def\eg{\emph{e.g}\onedot} 
\def\ie{\emph{i.e}\onedot}

\makeatother

\makeatletter

\def\OurMethod{VITA-Audio\xspace}

\def\OurTitle{\OurMethod: Fast Interleaved Cross-Modal \\ Token Generation for Efficient Large Speech-Language Model}

\makeatother

\definecolor{babyblue}{rgb}{0.54, 0.81, 0.94}
\definecolor{babyblueeyes}{rgb}{0.63, 0.79, 0.95}
\makeatother

\usepackage[normalem]{ulem}

\newcommand{\ZH}[1]{\begin{CJK}{UTF8}{gbsn}#1\end{CJK}}

\makeatletter

\let\titleold\title
\renewcommand{\title}[1]{\titleold{#1}\newcommand{\thetitle}{#1}}
\def\maketitlesupplementary
   {
   \newpage
       {
        \centering
        \Large
        \textbf{\thetitle}\\
        \vspace{0.5em}Supplementary Material \\
        \vspace{1.0em}
       }
   }

%% file: sec/0_abstract.tex
\begin{abstract}
With the growing requirement for natural human-computer interaction, speech-based systems receive increasing attention as speech is one of the most common forms of daily communication. However, the existing speech models still experience high latency when generating the first audio token during streaming, which poses a significant bottleneck for deployment. To address this issue, we propose VITA-Audio, an end-to-end large speech model with fast audio-text token generation. Specifically, we introduce a lightweight Multiple Cross-modal Token Prediction (MCTP) module that efficiently generates multiple audio tokens within a single model forward pass, which not only accelerates the inference but also significantly reduces the latency for generating the first audio in streaming scenarios. In addition, a four-stage progressive training strategy is explored to achieve model acceleration with minimal loss of speech quality. To our knowledge, VITA-Audio is the first multi-modal large language model capable of generating audio output during the first forward pass, enabling real-time conversational capabilities with minimal latency. VITA-Audio is \textbf{fully reproducible} and is trained on open-source data \textbf{only}. Experimental results demonstrate that our model achieves an inference speedup of \textbf{$3 \sim 5 \times$} at 7B parameter scale, but also significantly outperforms open-source models of similar model size on multiple benchmarks for automatic speech recognition~(ASR), text-to-speech~(TTS), and spoken question answering~(SQA).
\end{abstract}

%% file: sec/1_intro.tex
\section{Introduction}
\label{sec:1}

Real-time speech systems have become a crucial research focus for enabling natural dialogue.
Traditional speech systems predominantly adopt a modular design that decomposes real-time speech processing into three discrete components: automatic speech recognition (ASR), large language models (LLMs), and text-to-speech (TTS)~\cite{MThreads-full-duplex,AudioGPT,InternLM-XComposer2.5-OmniLive}.
However, this cascaded approach suffers from cumulative latency, loss of paralinguistic information (\eg, emotional prosody, rhythm) during modality conversion, and error accumulation between modules, substantially lowering the practical utility of cascaded architectures in real-time interactive scenarios.
To address the limitations of traditional methods, many recent studies have adopted an end-to-end approach to handle inputs and outputs of the model~\cite{MiniOmni,SLAM-Omni,LLaMA-Omni}.
These methods directly input speech into LLMs through an audio encoder and then synthesize speech response with discrete tokens~\cite{GLM-4-Voice} or LLM hidden states~\cite{Freeze-Omni}.

While existing end-to-end speech models generate output in a streaming fashion to reduce the response latency, their first token delay is still high.
Specifically, current speech models cannot directly deliver the first streaming audio chunk upon completing the first LLM forward pass, \ie, decoding the first text token.
In the applications requiring high real-time performance, this delay poses a significant bottleneck to the deployment of LLMs for speech processing.
This prompts a pertinent question:

\textit{How can we achieve more real-time audio generation within end-to-end speech models?} 

\begin{figure}[t]
\centering
\begin{subfigure}[b]{1\textwidth}
\centering
\includegraphics[width=0.97\textwidth]{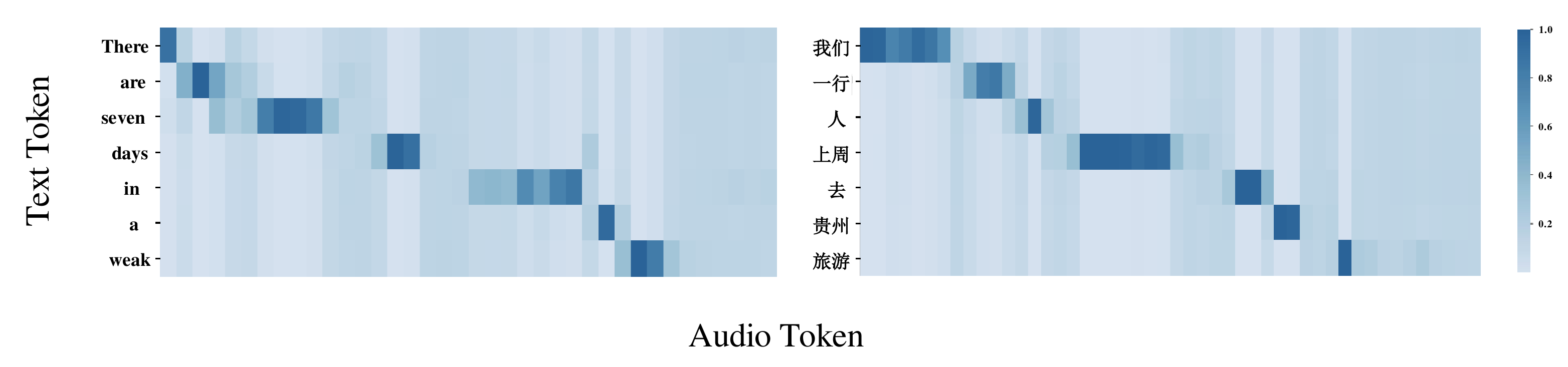}
\caption{The visualization of the attention maps between text tokens and audio tokens.}
\label{attn_weight_a}
\end{subfigure}
\vfill
\begin{subfigure}[b]{1\textwidth}
\centering
\includegraphics[width=\textwidth]
{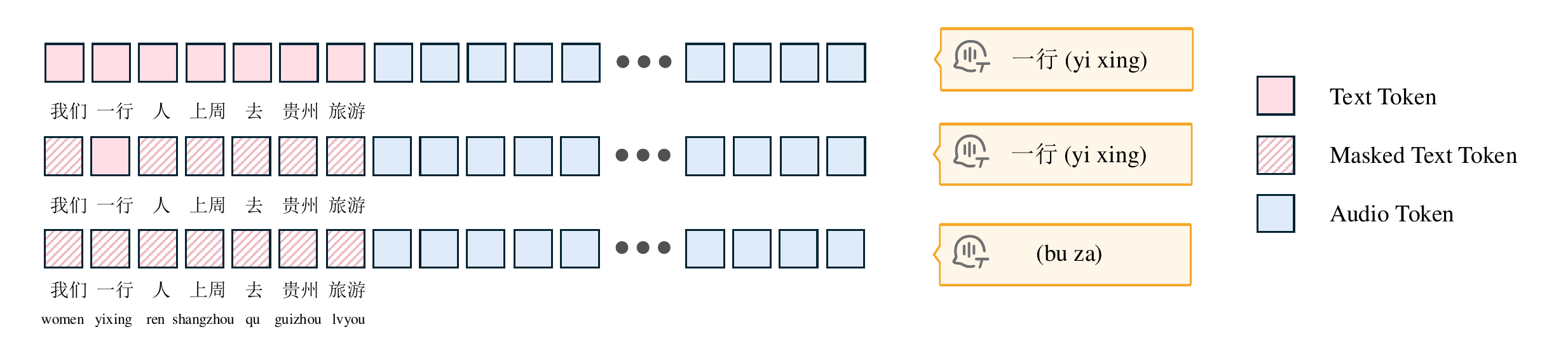}
\caption{The transcription results of the generated audio into text under different attention masks.}
\label{attn_weight_b}
\end{subfigure}
\caption{
(a) The audio sequence generated by the speech model exhibits a strong correlation with the corresponding text tokens.
(b) With irrelevant text tokens being masked out, the model is still able to generate the correct audio, and the pronunciation remains contextually appropriate.
However, if all text tokens are masked, the model outputs random audio.
This suggests that the hidden states from the LLM include sufficient contextual information for generating the corresponding audio tokens.
Consequently, the mapping from text hidden states to audio tokens is accomplished using relatively simple modules, without the need for the extensive semantic modeling typically required by LLMs.
}
\label{fig:attn_weights}
\end{figure}

To explore this issue, we visualized the hidden states of the final decoder layer of the speech model.
As shown in~\cref{attn_weight_a}, the audio tokens generated by the speech model show increased attention to the text tokens they correspond to.
As the generation of audio tokens progresses, the text tokens attended by the new audio token advance accordingly.
This finding is also reported in many literature on attention-based speech systems~\cite{LAS,Joint-CTC-attention}

In~\cref{attn_weight_b}, we show a Chinese sentence with a homograph ``\ZH{行}'' as an example.
The pronunciation of this character can be /xing/ or /hang/ depending on its context.
In Chinese, 'yihang' emphasizes spatial arrangement (e.g. a row/line of egrets), while 'yixing' focuses on the concept of a group or unit (e.g. a group/party of people).
Our speech model correctly decides the character's pronunciation to be the former, given the hidden states of historical inputs.
We then modify the model inference process by masking out all text hidden states, except the one corresponding to the token ``\ZH{一行}''(/yixing/) before generating its corresponding audio tokens. 
This modification prevents subsequently generated audio tokens from directly attending to other text tokens, although they can still attend to previously generated audio tokens.
We find that the subsequently generated audio tokens accurately produce the sounds as ``yixing'', which remains contextually appropriate.
The same observation also holds for other non-homograph tokens.
We therefore argue that \textbf{the hidden states from the LLM include sufficient contextual information for generating its corresponding audio tokens, and attending to additional texts is unnecessary}. 
Finally, we experiment with masking out all text tokens.
This time, the generated audio fails to align with its text and sounds like random non-speech babbles even though the model has access to the previously generated audio tokens.

These findings suggest that the speech model learns to primarily focus on the small span of corresponding text hidden states without heavily modeling the semantic space of the entire text and audio sequence.
This discovery instills confidence that we can learn the simple mapping relationship between text hidden states and audio tokens with relatively simple modules and without relying on the extensive semantic modeling of LLMs.

In this paper, we introduce \OurMethod, a lightweight framework that uses separate efficient modules, named Multiple Cross-modal Token Prediction (MCTP), to efficiently generate audio responses from text embeddings and LLM hidden states.
This approach enables obtaining both text tokens and an audio chunk in a single LLM forward pass, achieving zero delay in audio tokens.
A comparison of the delay of the first audio token is presented in~\cref{audio_token_delay}
, where we define ``audio token delay'' as the number of additional LLM forward steps required to generate the first audio token after the first LLM forward pass.
We distinguish this delay from ``audio generation delay'' which is the number of additional LLM forward passes to generate a meaningful and consistent chunk of audio.

\OurMethod has both zero audio token delay and zero audio generation delay.

To this end, through a four-stage progressive training strategy, we construct a set of lightweight yet powerful MCTP modules, which predict \textbf{10} audio tokens directly from historical inputs and LLM hidden states without requiring additional LLM forward passes, thus significantly enhancing the model's inference speed without sacrificing audio quality.

In summary, our main contributions are as follows.
\begin{itemize}
\item
We introduce \OurMethod, the first end-to-end speech model capable of generating audio during the first forward pass.
Leveraging audio generation without relying on extensive text semantic modeling capabilities, \OurMethod designs lightweight MCTP modules to generate decodable audio token chunks with zero audio token delay, thus overcoming the real-time limitations in traditional cascaded models and existing end-to-end methods.
\item
\OurMethod achieves remarkable end-to-end inference acceleration by generating ten audio tokens in a single forward pass, resulting in $3 \sim 5 \times$ speedup when implemented on a 7B LLM while preserving the ability of high-quality speech synthesis.
\item
We fully release \OurMethod to the open-source community.
Although \OurMethod is trained on open-source data only, comprehensive evaluations reveal that \OurMethod achieves the state-of-the-art performance on multiple benchmarks for ASR, TTS, and SQA tasks, outperforming existing models in both efficiency and accuracy, especially the open-source ones with a similar parameter scale, therefore setting a new standard for real-time speech-to-speech models.
\end{itemize}

\input{tables/audio_token_delay.tex}

%% file: tables/audio_token_delay.tex
\begin{table*}[t!]
\caption{Comparison of recent speech models, \OurMethod leverages the hidden state to enhance model performance, adopts an end-to-end architecture, and achieves zero audio token delay.}
\begin{center}
\begin{adjustbox}{max width=0.80\textwidth}
\begin{tabular}{lccc}
\toprule
{Model} & {Audio Token Delay} & {Leveraging Hidden States} & {End-to-End} \\
\midrule
Freeze-Omni~\hfilll~\cite{Freeze-Omni} & Text Length & \Checkmark & \XSolidBrush \\
MinMo~\hfilll~\cite{MinMo} & 5 & \Checkmark & \XSolidBrush \\
Mini-Omni~\hfilll~\cite{MiniOmni} & 7 & \XSolidBrush & \Checkmark \\
Moshi~\hfilll~\cite{Moshi} & 1 & \XSolidBrush & \Checkmark \\
GLM-4-Voice~\hfilll~\cite{GLM-4-Voice} & 13 & \XSolidBrush & \Checkmark \\
LUCY~\hfilll~\cite{LUCY} & 7 & \XSolidBrush & \Checkmark \\
\midrule
\OurMethod & 0 & \Checkmark & \Checkmark \\
\bottomrule
\end{tabular}
\end{adjustbox}
\end{center}
\label{audio_token_delay}

\end{table*}

%% file: sec/2_related.tex
\section{Related Work}

Large language models (LLMs) have revolutionized human-computer interaction with advanced natural language processing. Extending these capabilities to speech—a natural communication modality—has become a key research focus. Traditional speech interaction systems~\cite{MThreads-full-duplex,AudioGPT,InternLM-XComposer2.5-OmniLive} typically adopt a cascaded architecture, combining separate ASR, LLM, and TTS modules. However, this approach suffers from increased latency, loss of paralinguistic cues, and error propagation.

Recent work~\cite{VITA,Step-Audio} has improved integration by connecting audio encoders to LLMs via trainable adapters, but still relies on independent TTS modules. To address this, some methods incorporate LLM hidden states into audio decoders. For example, Llama-Omni~\cite{LLaMA-Omni} uses a non-autoregressive transformer to predict audio tokens from upsampled LLM states, while Freeze-omni~\cite{Freeze-Omni} freezes the LLM and combines autoregressive and non-autoregressive decoders. Minmo~\cite{MinMo} integrates a language model with CosyVoice2~\cite{CosyVoice2} for mixed speech-text processing.

End-to-end models further unify TTS within LLMs, enabling direct text and speech generation. These models follow either parallel or interleaved audio-text modeling. 
In the parallel modeling paradigm, the model uses different heads to process hidden states, generating both text and multiple audio tokens~\cite{SLAM-Omni,Moshi}.
Since the input to the LLM is altered during autoregression, maintaining the original capabilities of the LLM presents significant challenges.
To perform inference without large-scale pretraining, Mini-Omni~\cite{MiniOmni} and LUCY~\cite{LUCY} rely on batch parallel decoding to preserve the inference capability of the LLM.

Compared to parallel-paradigm models, interleave-paradigm models appear to better preserve language capabilities, as suggested by the performance comparison on spoken question-answering benchmarks~\cite{GLM-4-Voice}.
We attribute this difference to the fact that parallel-paradigm models use an average of text and audio representations as input, which significantly diverges from the inputs used during pretraining.
However, interleave-paradigm models face a latency issue due to their sequential prediction of audio tokens, especially when the audio token rate is high.

\OurMethod leverages the strengths of these architectures by adopting the interleaved modeling paradigm and introducing MCTP for audio generation.
The former maximally preserves the LLM's language ability, and the latter reduces inference latency by generating multiple audio tokens in a single forward pass.

%% file: sec/3_method.tex
\section{Method}
\label{sec:3}

\begin{figure}[t]
\centering
\includegraphics[width=0.95\textwidth,]{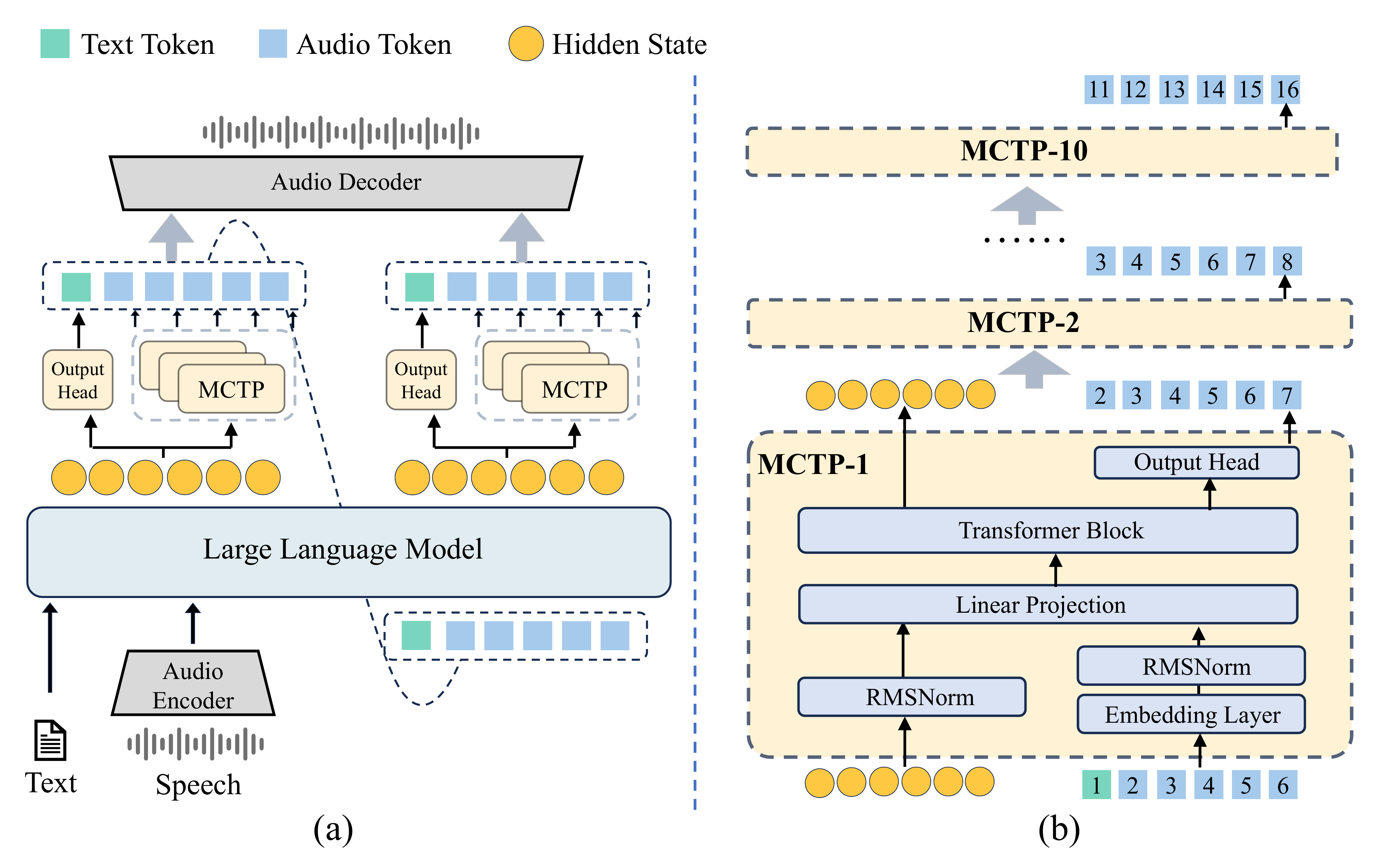}
\caption{
Architecture overview. (a) \OurMethod is an end-to-end large speech model equipped with 10 light-weight Multiple Cross-modal Token Prediction (MCTP) modules that enable speech generation with extremely low latency. As shown in \cref{fig:attn_weights}, we observe that the hidden states of certain text tokens in the LLM backbone contain sufficient semantic information for generating the corresponding audio tokens, which means that it is unnecessary to attend to additional text tokens when generating audio. Thus, we propose to utilize a set of light-weight MCTP modules to model the mapping from LLM hidden states to the audio tokens. (b) The details of the MCTP modules. Our MCTP module has a light-weight architecture, which enables it to finish one forward pass within $0.0024$ seconds ( $11$\% of the LLM backbone). The MCTP module is capable of generating 10 audio tokens from the LLM hidden states and the text embedding, and the generated audio tokens can be decoded by the audio decoder directly. The utilization of MCTP modules enables \OurMethod to generate audio responses in one LLM forward pass, which achieves extremely fast generation speed. 
}
\label{arch}
\end{figure}
\subsection{Overview}

As illustrated in~\cref{arch}, \OurMethod consists of four major components: an audio encoder, an audio decoder, a large language model backbone, and a set of Cross-modal Token Prediction (MCTP) modules.
The audio signal is first processed by the audio encoder, whose output is then fed into the LLM for further processing.
During each forward pass, the LLM alternately generates text and audio tokens.
The hidden states from the final layer of LLM, along with the embedding of the predicted token, are provided as input to the MCTP modules.
The historical input tokens, the tokens predicted by the LLM, and by the MCTP modules are concatenated to form the inputs to the next LLM forward pass.
Finally, the audio tokens generated by both the LLM and the MCTP modules are aggregated and passed to the audio decoder to generate the final audio output.

\subsection{Multiple Cross-Modal Token Prediction (MCTP) Module}
As described in~\cref{sec:1}, the text and speech modalities exhibit a monotonic alignment pattern.
This cross-modal alignment allows us to avoid complex modeling of the semantic latent space and to focus on learning a simple text-to-speech mapping relationship, which we propose to use lightweight modules to learn.
In the preliminary experiments, we use a few lightweight Transformer blocks to predict multiple audio tokens from LLM hidden states, and embed the predicted tokens into the LLM's autoregressive inference.

Standard autoregressive modeling can be formulated as:
\begin{equation}
	p_t(Y_{t-1},...,Y_0) \equiv P[Y_t|Y_{t-1},...,Y_0],
\end{equation}
where $Y_t$ denotes the predicted audio token at time step $t$, and $p_t$ represents the conditional probability distribution based on the historical sequence.
When extended to multi-step prediction, \ie, predicting the $i\text{-}th$ audio token at time step $t+i$, the formulation becomes as:
\begin{equation}
	p_{t+i}(Y_{t-1},...,Y_0) \equiv \widetilde{P}[Y_{t+i}|Y_{t-1},...,Y_0].
\end{equation}
At this point, there is a significant deviation in the consistency of the distribution between $\widetilde{P}$ and $P$.
As $i$ increases, the difference between the two distributions will progressively widen, resulting in a growing accumulation of errors and leading to poor mapping between text and audio.

To address this issue, we adopt a cascaded prediction architecture.
Specifically, the hidden states and output sequence from the preceding modules are employed as joint input conditions for the subsequent modules:
\begin{equation}
	p_{t+i}(Y_{t-1}, \dots, Y_0) \equiv \widetilde{P}[Y_{t+i} | Y_{t-1}, \dots, Y_0, h_{t+i-1}, o_{t+i-1},\dots,o_{t}]
	,
\end{equation}
where $h_{t+i-1}$ and $o_{t+i-1}$ represent the hidden state and output sequence of the preceding module, respectively.
By introducing progressively updated contextual information, modules can achieve incremental optimization of cross-modal mapping, ensuring accurate modality synchronization at each time step.
Inspired by DeepSeek V3~\cite{DeepSeek-V3}, we adopt an isomorphic Multi-Token Prediction (MTP) framework to construct our MCTP module.
Unlike DeepSeekV3 and speculative decoding~\cite{EAGLE}, where the former focuses on improving training and the latter requires verification to ensure that the sampling distribution matches exactly with that of the original model, we use the MCTP module for audio-text mapping, presumably a simpler task than semantic modeling.
As a result, we require a comparatively small amount of text data to train our model.

Since the embedding layer and output heads are shared with the LLM, the audio tokens generated by the MCTP module are directly incorporated into the autoregressive process of the LLM.
As illustrated in~\cref{arch}, the hidden states and output token, from the LLM or the preceding MCTP module, are concatenated with the input tokens and fed into a Transformer block for next-step processing.
The resulting hidden states and token are then passed to the next MCTP module.
Upon completion of a forward pass, the audio tokens generated by either the LLM or the MCTP modules are aggregated as the input sequence for the subsequent LLM forward iteration.

\subsection{Training}
\subsubsection{Data Construction}

\OurMethod is trained exclusively on open-source datasets, integrating multi-domain and multi-language speech data resources.
The training dataset encompasses a diverse range of sources.
Detailed descriptions of the datasets used at each stage are provided in~\cref{table_data} in the Appendix.

All training data are uniformly packed into sequences of fixed length (8K tokens), an approach that enables effective training on samples of varying lengths~\cite{Long-VITA}.
We reinitialize the positional embeddings and attention masks for all packed samples to ensure that the model attends exclusively to tokens within the same original sample.
This processing strategy not only eliminates potential artifacts introduced by data concatenation but also significantly enhances training stability and reduces computational overhead.

\subsubsection{Training Pipline}

\input{figures/VITA_audio_training.tex}
For \OurMethod to output a consistent sequence of audio tokens in a single forward pass, each MCTP module must model a distinct distribution.
As a result, training all the MCTP modules simultaneously becomes a challenging task, especially when the number of modules is large, due to potentially misaligned optimization objectives.
We propose a four-stage training strategy, as shown in~\cref{training_pipline}, to progressively equip the MCTP modules with the ability to map text to its audio, thereby reducing the difficulty of their convergence.
Further training details are provided in Sec.~\ref{training_append} of the Appendix.

\input{figures/mode}

\subsection{Inference}
\label{inference}

In order to address diverse scenarios, four distinct inference paradigms have been designed as shown in~\cref{fig:mode}.

For the ASR and TTS tasks, we propose \OurMethod-Turbo.
In each forward pass, the LLM generates one token, followed by the generation of ten tokens by the MCTP modules.
This paradigm is the most efficient among the options; however, the performance of speech dialogue tasks degrades due to the need to predict the text token.
For speech dialogue tasks, we introduce \OurMethod-Boost and \OurMethod-Balance.
Their main difference lies in audio token generation: \OurMethod-Boost generates eight directly decodable audio tokens in the first forward pass, whereas \OurMethod-Balance adheres to a strict $1:2$ text-to-audio token ratio for enhanced speech quality.
The latter requires two forward passes to generate enough audio tokens for decoding.
To optimize model performance, distinct models were trained for each of the two modes.

\OurMethod-Vanilla is designed for scenarios that require higher language performance.
It generates tokens solely using the main LLM, sacrificing efficiency but offering a slight performance boost.

%% file: figures/VITA_audio_training.tex
\begin{figure}[t]
\centering
\includegraphics[width=0.95\textwidth]{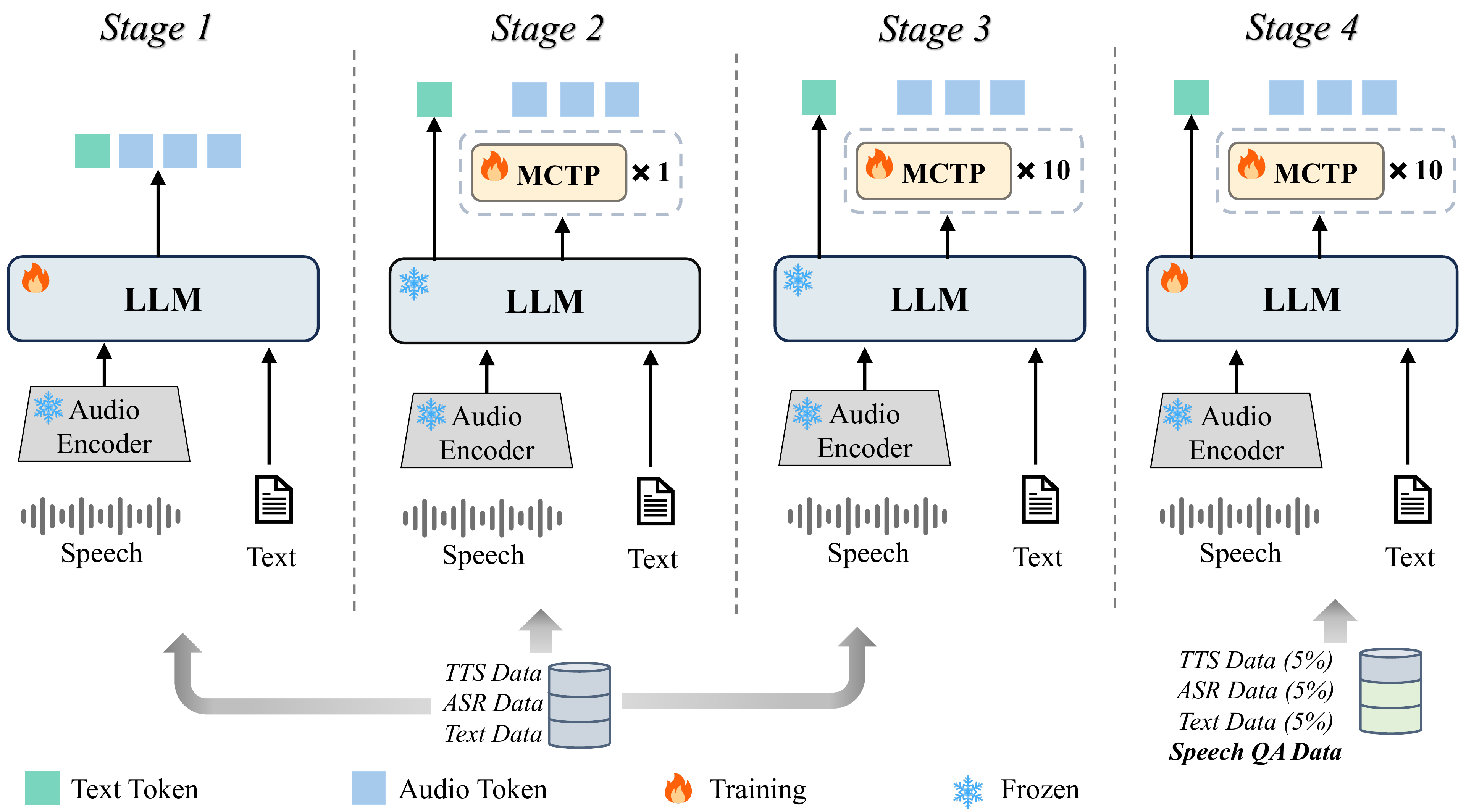}
\caption{
{Training pipeline of \OurMethod.}
The first stage (Audio-Text Alignment) enhances the LLM by extending its audio modeling capability through large-scale speech pre-training.
The second stage (Single MCTP module Training) connects an MCTP module with the LLM to predict one subsequent token based on the input tokens and the LLM's hidden states.
The third stage (Multiple MCTP Modules Training) increases the number of MCTP modules in the model to predict more tokens in each model forward.
The last stage (Supervised Fine-tuning) provides the speech-to-speech capability to the model by optimizing it on the large-scale speech QA dataset.
}
\label{training_pipline}
\end{figure}

%% file: figures/mode.tex
\begin{figure}[t]
\centering
\includegraphics[width=1\textwidth]{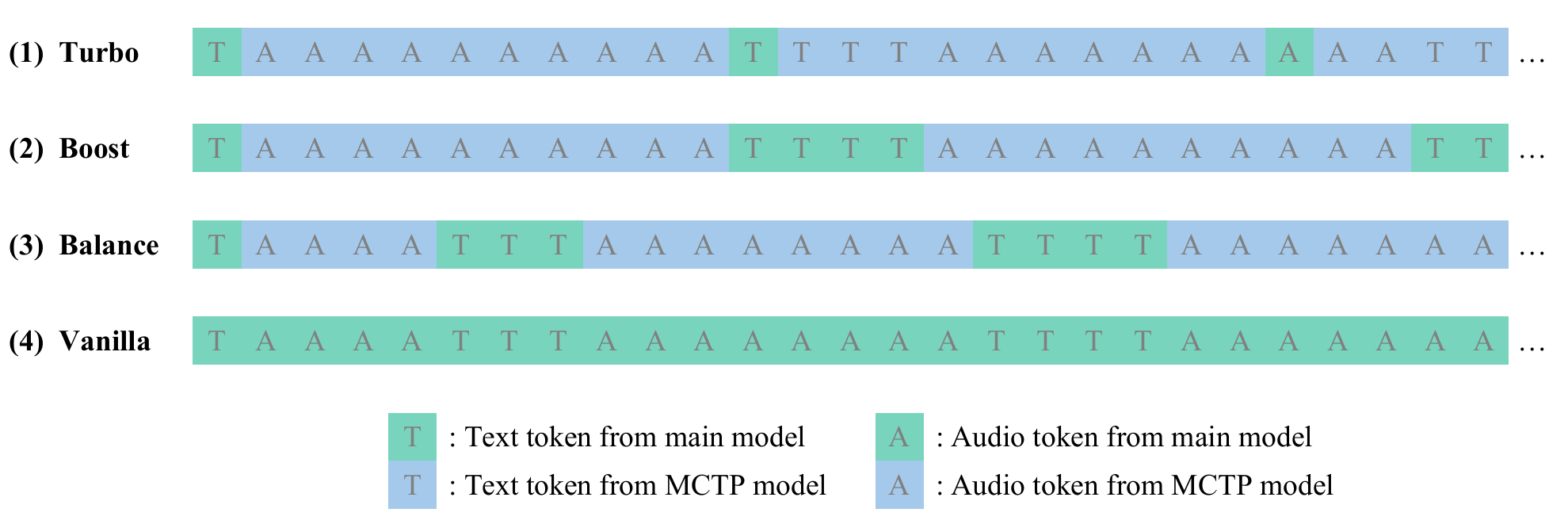}
\caption{
The four text-audio interleaved inference modes are illustrated as follows:
1) Turbo:
As the fastest inference mode, it generates $1$ token by the main model and $10$ additional tokens via MCTP in each forward pass.
To ensure that a valid audio chunk is decoded after the first forward pass, the first generated $11$ tokens are split into $1$ text token and $10$ audio tokens.
Then, the Turbo mode iteratively generates $4$ text tokens and $10$ audio tokens in the following forward.
2) Boost:
To enhance the quality of text tokens, Boost mode follows the text-audio cyclic pattern of Turbo mode, with the main model generating every text token and MCTP generating every audio token.
3) Balance:
To keep a balanced text-audio ratio, \ie, $1:2$, the balance mode further changes the text-audio cyclic pattern of the Boost mode.
Specifically, the balance mode sequentially generates $1$ text token from the main model, $4$ audio tokens ($2$ tag tokens mark the beginning and end of audios, and $2$ common tokens denote the audio content) from MCTP, $3$ text tokens from the main model, $8$ text tokens ($2$ tag tokens mark the beginning and end of audios, and $6$ common tokens denote the audio content) from MCTP, and then iteratively generates $4$ text tokens from the main model and $10$ audio tokens ($2$ tag tokens mark the beginning and end of audios, and $8$ common tokens denote the audio content) from MCTP.
4) Vanilla:
As the slowest inference mode, Vanilla mode follows the text-audio cyclic pattern of Balance mode, with the main model generating every token.
}
\label{fig:mode}
\end{figure}

%% file: sec/4_experiment.tex
\section{Experiment}
\label{sec:4}

\subsection{Experiment Settings}

We use the Qwen2.5-7B-Instruct~\cite{Qwen2.5} as the
pre-trained text LLM.
The initial version of \OurMethod utilizes the speech tokenizer and speech decoder in GLM-4-Voice~\cite{GLM-4-Voice}, which effectively captures semantic information at an ultra-low bitrate.
In the second version, \ie, \OurMethod-Plus further replaces the GLM-4-Voice tokenizer with SenseVoiceSmall~\cite{FunAudioLLM} and an MLP-based adapter.
The detail comparison between \OurMethod and \OurMethod-Plus is listed in~\cref{table_model}.

\input{tables/spoken_question_answering.tex}

\input{tables/text_to_speech.tex}

\input{tables/asr_main}

\subsection{Evaluation on Spoken Question Answering}

We evaluate the spoken question answering capability of \OurMethod on three public English datasets: Web-Questions~\cite{Web-Questions}, Llama-Question~\cite{Llama-Question}, and TriviaQA~\cite{TriviaQA}.
Two evaluation methods are employed:
S$\rightarrow$T, where the text responses generated by the model are evaluated directly, and S$\rightarrow$S, where the model’s speech responses are transcribed using Whisper~\cite{Whisper} before evaluation.
We compare \OurMethod with the latest speech models that have comparable parameter sizes, and the results are shown in~\cref{table_SQA}.
Our model demonstrates superior performance in the S$\rightarrow$T task and achieves SOTA results in the S$\rightarrow$S setup.
It is particularly noteworthy that the training approach of \OurMethod ensures minimal degradation between S$\rightarrow$T and S$\rightarrow$S, with a performance drop of only $9\%$.
This indicates that \OurMethod achieves high-quality alignment between text and speech modalities, with benefits extending beyond processing speed alone.

\subsection{Evaluation on Fundamental Speech Competence}

\textbf{TTS}\quad
We evaluate the TTS performance of \OurMethod on Seed-TTS~\cite{Seed-TTS} and LibriTTS~\cite{LibriTTS} benchmarks.
We use Whisper-Large-V3~\cite{Whisper} and Paraformer\cite{Paraformer} to transcribe into text the generated English and Chinese speech, respectively.

We present the results of \OurMethod at each stage in~\cref{table_TTS}. 
In these results, the output of \OurMethod's first stage (S1) consists of text tokens directly generated by the LLM, while the outputs of the second (S2), third (S3), and fourth (S4) stages are alternately generated by both the LLM and the MCTP module.
The experiments demonstrate that \OurMethod outperforms other open-source models with a similar number of parameters.
Additionally, it should be noted that, despite using ten MCTP modules for accelerated inference, \OurMethod's TTS capabilities are largely preserved throughout the training process, further validating the effectiveness of the MCTP module in aligning text and audio.

\textbf{ASR}\quad
We evaluate the ASR performance of the four stages of \OurMethod on WenetSpeech~\cite{WenetSpeech}, AIshell~\cite{AISHELL_1}, LibriSpeech~\cite{Librispeech}, and Fleurs~\cite{Fleurs}, and a subset of the results are reported in~\cref{tab:ASR}. More detailed results can be found in~\cref{table_ASR_1} and~\cref{table_ASR_2} in the Appendix.
The results for other works are partially reproduced from their respective original works for comparison.

It can be observed that \OurMethod-plus-vanilla demonstrates highly competitive performance across various benchmarks. Moreover, \OurMethod-plus-Boost achieves remarkably fast inference speed while still maintaining strong overall performance.

\input{tables/speed}

\subsection{Evaluation of Latency}
\textbf{Inference Speedup}\quad
Efficient mapping between text and speech is the core of \OurMethod.
To demonstrate its effectiveness, we compare the inference time across different modes of \OurMethod for various model sizes.
Specifically, we evaluate the inference speed on GPUs capable of 148 TFLOPS under bfloat16 precision, with the output fixed at $4096$ tokens, and record the total time as the model’s inference time.
All models, regardless of size, are randomly initialized, and this initialization do not affect the inference time measurements.
We use Transformers~\cite{Transformers} and FlashAttention-2~\cite{FlashAttention-2}.
As mentioned in~\cref{inference}, \OurMethod-Vanilla only uses the main model for output; \OurMethod-Turbo uses both the main model and all the MCTP modules during each forward pass; and \OurMethod-Boost and \OurMethod-Balance progressively increase the number of MCTP modules used to ensure higher accuracy.
As shown in~\cref{tab:inference_comparison}, we present a comparison of the time consumption for different inference modes and model sizes.
Noted that the time consumption does not include the cost of the audio encoder and audio decoder.
We observe that in \OurMethod-Turbo, a speedup of approximately $5\times$ is achieved across models ranging from $0.5$B to $72$B, greatly enhancing the output token throughput.
\OurMethod-Boost also achieves around $3\times$ acceleration across various sizes, resulting in a desirable performance for real-time speech dialogue systems.
For example, $72$ B \OurMethod generates approximately $40$ tokens per second, which, excluding generated text tokens, corresponds to roughly three seconds of audio and associated text when using a $12.5$Hz speech tokenizer.
This performance is sufficiently fast for human-computer interaction.

\textbf{Latency}\quad
In human-computer interaction, latency is a crucial metric, as it determines whether users can interact with the model in real-time.
Given that most speech models support streaming output, the key to reducing perceived latency lies in shortening the time required to generate the first chunk of audio.

\input{figures/timeline}

We visualize the timeline of model decoding phrase in~\cref{fig:timeline}.
The green marks denote the tokens generated by the main model, and the blue marks are the tokens generated by MCTP modules.
We set the number of prefiil tokens to $32$.
And~\cref{fig:timeline} shows that \OurMethod-Turbo completes the generation of the first audio chunk in about $50$ ms, while \OurMethod-Vanilla requires about $220$ ms.
\OurMethod-Boost and \OurMethod-Balance generate fewer audio tokens in the first forward and more text audio tokens in the following forward.
Thus, they are slower than \OurMethod-Turbo but still significantly faster than \OurMethod-Vanilla.
Thanks to the advantage of zero audio generation delay, \OurMethod produces multiple audio tokens in the first forward pass, allowing the first audio token chunk to be generated during the initial forward pass, which can then be used for decoding.
This significantly reduces the perceived delay.
In the experimental environment previously mentioned, \OurMethod reduces the time to generate the first audio token chunk from $236$ to $53$ ms, as shown in~\cref{e2e_time}.

%% file: tables/spoken_question_answering.tex
\begin{table*}[t]
\caption{
Results on Spoken Question Answering~(SQA) benchmarks.
``S$x$'' denotes the $x$-th training stage of speech models.
}
\label{table_SQA}

\begin{center}
\begin{adjustbox}{max width=0.99\textwidth}
\begin{tabular}{lc|cc|cc|cc|cc}
\toprule

\multirow{2}{*}{Model} 
& \multirow{2}{*}{\#Params} 
& \multicolumn{2}{c|}{Llama Question}
& \multicolumn{2}{c|}{TriviaQA}
& \multicolumn{2}{c|}{Web Question} 
& \multicolumn{2}{c}{Mean} \\
& & S $\rightarrow$ T & S $\rightarrow$ S & S $\rightarrow$ T & S $\rightarrow$ S & S $\rightarrow$ T & S $\rightarrow$ S & S
$\rightarrow$ T & S $\rightarrow$ S \\

\midrule
\multicolumn{10}{c}{Proprietary Models} \\

MinMo~\hfilll~\cite{MinMo} & 7B & 78.9 & 64.1 & 48.3 & 37.5 & 55.0 & 39.9 & 60.7 & 47.2\\

\midrule
\multicolumn{10}{c}{Open-source Models} \\

Moshi~\hfilll~\cite{Moshi} & 7B & 62.3 & 21.0 & 22.8 & 7.3 & 26.6 & 9.2& 37.2 & 12.5 \\

GLM-4-Voice~\hfilll~\cite{GLM-4-Voice} & 9B & 64.7 & 50.7 & {39.1} & 26.5 & \textbf{55.0} & {39.9} & 45.3 & 31.0 \\

LUCY (S2)~\hfilll~\cite{LUCY} & 7B & 59.6 & 51.0 & 23.2 & 18.2 & 26.6 & 18.2 & 36.5 & 29.1 \\

\midrule
\OurMethod-Boost & 7B & 68.7 & 60.3 & 30.5 & 29.3 & 32.9 & 30.4 & 44.0 & 40.0 \\

\OurMethod-Vanilla & 7B & 71.3 & 66.3 & 31.9 & 30.1 & 33.5 & 31.4 & 45.6 & 42.6 \\

\OurMethod-Plus-Boost & 7B & \textbf{76.3} & 64.6 & 43.6 & 39.5 & 44.2 & 40.0 & 54.7 & 48.0 \\

\OurMethod-Plus-Vanilla & 7B & 75.6 & \textbf{68.0} & \textbf{45.9} & \textbf{42.7} & 45.0 & \textbf{41.7} & \textbf{55.5} & \textbf{50.8} \\

\bottomrule
\end{tabular}
\end{adjustbox}
\end{center}

\end{table*}

%% file: tables/text_to_speech.tex
\begin{table*}[t]
\caption{Results on Text to Speech~(TTS) Benchmarks.
``S$x$'' denotes the $x$-th training stage.
}
\label{table_TTS}

\centering
\begin{adjustbox}{max width=0.99\textwidth}
\begin{tabular}{l|ccc|c}
\toprule
\multirow{3}{*}{Model} & \multicolumn{3}{c}{Seed-TTS} & \multicolumn{1}{c}{LibriTTS} \\

& \textit{test-zh} & \textit{test-en} & \textit{test-hard} & \textit{test-clean} \\
& CER (\%)~$\downarrow$ & WER (\%)~$\downarrow$ & WER (\%)~$\downarrow$ & WER (\%)~$\downarrow$ \\

\midrule
Seed-TTS~\hfilll~\cite{Seed-TTS} & 1.12 & 2.25 & 7.59 & -- \\

CosyVoice~\hfilll~\cite{CosyVoice} & 3.63 & 4.29 & 11.75 & 2.89 \\
CosyVoice2~\hfilll~\cite{CosyVoice2} & 1.45 & 2.57 & \textbf{6.83} & 2.47 \\

VITA-1.5 (S3)~\hfilll~\cite{VITA-1.5} & 8.44 & 2.63 & -- & -- \\
GLM-4-Voice~\hfilll~\cite{GLM-4-Voice} & 2.91 & 2.10 & -- & 5.64 \\

\midrule

\OurMethod-Turbo (S1) & 1.18 & 1.92 & 10.58 & 1.96 \\
\OurMethod-Turbo (S2) & \textbf{0.96} & 1.92 & 9.72 & 1.98 \\
\OurMethod-Turbo (S3) & 1.05 & \textbf{1.77} & 9.86 & 1.99 \\
\OurMethod-Turbo (S4) & 1.07 & 2.26 & 10.08 & 2.08 \\

\OurMethod-Plus-Boost & 1.32 & 2.21 & 12.05 & 2.21 \\
\OurMethod-Plus-Vanilla & 1.13 & 1.85 & 10.21 & \textbf{1.89} \\

\bottomrule
\end{tabular}
\end{adjustbox}
\end{table*}

%% file: tables/asr_main.tex
\begin{table*}[t]
\caption{Results on Automatic Speech Recognition~(ASR) Benchmarks.
``S$x$'' denotes the $x$-th training stage.
Compared to other methods, \textbf{\OurMethod is trained with open-source data only}.
}

	\label{tab:ASR}
	\begin{center}
		\begin{adjustbox}{max width=0.99\textwidth}
			\begin{tabular}{lccccc}
				\toprule
				\multirow{2}{*}{{Model}} & \multicolumn{2}{c}{{WenetSpeech}} & {AIShell} & \multicolumn{2}{c}{{LibriSpeech}} \\\cmidrule(lr){2-3}\cmidrule(lr){4-4}\cmidrule(lr){5-6}
				& test\_meeting~$\downarrow$ & test\_net~$\downarrow$ & test~$\downarrow$ & test-clean~$\downarrow$ & test-other~$\downarrow$ \\
				\midrule
				Qwen2-Audio-base~\hfilll~\cite{Qwen2-Audio} & 8.40 & 7.64 & 1.52  & \textbf{1.74} & \textbf{4.04} \\
                Baichuan-Audio-base~\hfilll~\cite{Baichuan-Audio}& 13.28 & 10.13 & 1.93 & 3.02 & 6.04 \\
				VITA-1.5 (S3)~\hfilll~\cite{VITA}  & 10.0 &8.4  & 2.2  & 3.4 & 7.5  \\
				Freeze-Omni~\hfilll~\cite{Freeze-Omni}  & 13.46 & 11.8 & 2.48 & 3.82 & 9.79 \\
				LUCY (S1)~\hfilll~\cite{LUCY}  & 10.42 & 8.78 & 2.40 & 3.36 & 8.05 \\
                Step-Audio-chat~\hfilll~\cite{Step-Audio} & 10.83 & 9.47  & 2.14 & 3.19 & 10.67 \\
                Qwen2.5-Omni~\hfilll~\cite{Qwen2.5-Omni} & 7.71 & \textbf{6.04}  & \textbf{1.13} & 2.37 & 4.21 \\

				\midrule

				\OurMethod-Vanilla  & 17.34 & 13.45 & 4.46 & 2.98 & 8.07 \\
				\OurMethod-Plus-Boost  & 9.38 & 8.97 & 4.72  & 3.13 & 7.07 \\
				\OurMethod-Plus-Vanilla(S1)  & \textbf{6.68} & 6.59 & 1.51  & 1.91 & 4.29 \\
				\OurMethod-Plus-Vanilla  & 7.12 & 6.90 & 1.94 & 2.00 & 4.60 
                \\

				\bottomrule

			\end{tabular}
		\end{adjustbox}
		
	\end{center}
\end{table*}

%% file: tables/speed.tex
\begin{table*}[t]
\caption{Boostup Ratio under Different Inference Paradigms.}
\label{tab:inference_comparison}
\begin{center}
\begin{tabular}{@{\hskip 5pt} l @{\hskip 10pt} c @{\hskip 10pt} c @{\hskip 10pt} c @{\hskip 10pt} c @{\hskip 5pt} c @{\hskip 5pt}}
\toprule
{Mode} & {{Model Size}} & {{\#GPU}} & {Total Second $\downarrow$} & {Token Per Second $\downarrow$} & {Speedup $\uparrow$} \\
\midrule
Vanilla& \multirow{4}{*}{0.5B} & \multirow{4}{*}{1} & 53.89 & 76.00 & 1.00 $\times$ \\
Boost& & & 20.65& 198.35 &  2.61 $\times$ \\
Balance & & & 20.71 & 197.78 & 2.60 $\times$ \\
Turbo & & & 11.83 & 346.24 & 4.56 $\times$ \\
\midrule
Vanilla & \multirow{4}{*}{7B} & \multirow{4}{*}{1} & 63.38 & 64.62 & 1.00 $\times$ \\
Boost & & & 23.97 & 170.88 & 2.64 $\times$ \\
Balance & & & 23.94 & 171.09 & 2.64 $\times$ \\
Turbo & & & 13.43 & 304.99 & 4.72 $\times$ \\
\midrule
Vanilla& \multirow{4}{*}{72B} & \multirow{4}{*}{2} & 255.13 & 16.05 & 1.00 $\times$ \\
Boost & & & 84.98 & 48.20 & 3.00 $\times$ \\
Balance & & & 85.13 & 48.11& 3.00 $\times$ \\
Turbo & & & 39.5 & 103.60 & 6.46 $\times$ \\
\bottomrule
\end{tabular}
\end{center}
\end{table*}

%% file: figures/timeline.tex
\begin{figure}[t]
	\centering
    \includegraphics[width=1.0\textwidth]{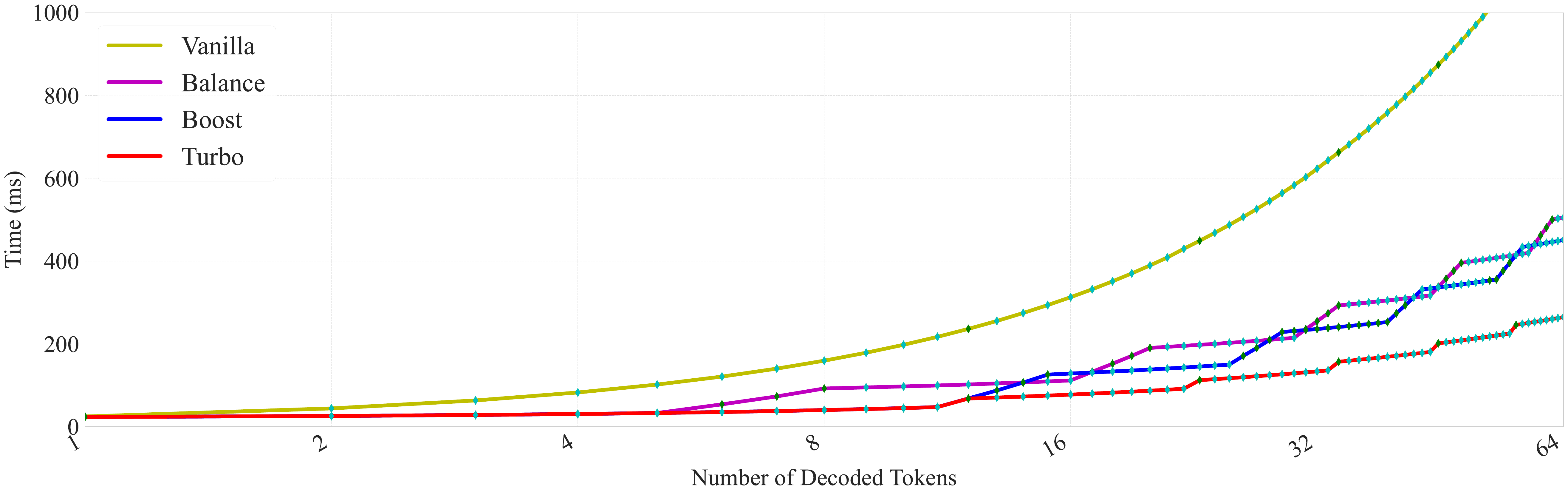}	
	\caption{
		 Token generation speed curves of four text-audio interleaved modes.
	}
	\label{fig:timeline}
\end{figure}

%% file: sec/5_conclusion.tex
\section{Conclusion}
In this paper, we introduce \OurMethod, a lightweight framework that uses separate efficient modules, named Multiple Cross-modal Token Prediction (MCTP) modules, to efficiently generate audio responses from text embeddings and LLM hidden states.
MCTP learns the simple mapping relationship between text hidden states and audio tokens with relatively simple modules and without relying on the extensive semantic modeling of LLMs.
Our model achieves new state-of-the-art performance on multiple benchmarks for ASR, TTS, and SQA tasks, outperforming existing models in efficiency and accuracy, especially the open-source ones with a similar parameter scale.
Therefore, it sets a new standard for real-time speech-to-speech models.

%% file: sec/append.tex
\appendix

\newpage

\setcounter{page}{1}
\maketitlesupplementary
\renewcommand\thefigure{\Alph{section}\arabic{figure}}
\renewcommand\thetable{\Alph{section}\arabic{table}}

\setcounter{figure}{0}
\setcounter{table}{0}

\section{Training Data} 
\label{da ta_append}

\textbf{ASR Data}\quad
We aggregated approximately $100,000$ hours of open-source Automatic Speech Recognition (ASR) data, including WenetSpeech~\cite{WenetSpeech}, Librispeech~\cite{Librispeech}, Multilingual LibriSpeech~\cite{MultilingualLibriSpeech}, Common Voice 17~\cite{Librispeech}, MMCRSC~\cite{MMCRSC}, GigaSpeech~\cite{GigaSpeech}, People's Speech~\cite{PeoplesSpeech}, VoxPopuli~\cite{VoxPopuli}, and the AISHELL series (AISHELL-1~\cite{AISHELL_1} to AISHELL-4~\cite{AISHELL_4}).

\textbf{TTS Data}\quad
Concurrently, we integrate  approximately $100,000$ hours of open-source Text-to-Speech (TTS) data, primarily consisting of the Wenetspeech4TTS~\cite{Wenetspeech4TTS}, LibriTTS~\cite{LibriTTS}, GLOBE~\cite{GLOBE}, and Emilia~\cite{Emilia} datasets.

\textbf{Speech QA Data}\quad For speech question-answering (Speech-QA),
we utilize VoiceAssistant-400K~\cite{MiniOmni} and AudioQA-1.0M~\cite{LUCY}, totaling $1.4$ million speech QA data, to enhance the model's speech-to-speech dialogue capabilities.

\textbf{Text-Only Data}\quad The pure text data is collected from OpenHermes-2.5~\cite{OpenHermes-2.5} LIMA~\cite{LIMA}, databricks-dolly-15k~\cite{databricks-dolly-15k}, MetaMathQA ~\cite{MetaMathQA}, MathInstruct~\cite{MathInstruct}, Orca-Math~\cite{Orca-Math}, atlas-math-sets~\cite{atlas-math-sets}, goat~\cite{goat}, and camel-ai-math~\cite{CAMEL}.
Given that discrete audio token sequences exhibit significantly longer lengths compared to their textual counterparts, we incorporate several specialized long-context text datasets following the Long-VITA~\cite{Long-VITA} to enhance contextual modeling capabilities.
These include Long-Instruction~\cite{Long-Instruction-with-Paraphrasing}, LongForm~\cite{LongForm}, LongAlign-10k~\cite{LongAlign}, LongCite-45k~\cite{LongCite}, LongWriter-6k~\cite{LongWriter}, LongQLoRA~\cite{LongQLoRA}, LongAlpaca~\cite{LongLoRA}, and LongData-Collections~\cite{Long-Data-Collections}.

\section{Training Pipline} 
\label{training_append}

\textbf{Stage 1: Audio-Text Alignment.}
Building upon pretrained language models, the goal of this stage is to extend the audio modeling capabilities to the LLM through large-scale speech pretraining.
We freeze the audio encoder and audio decoder, and train the LLM using ASR, TTS, and Text-only data. During this stage, the output of the LLM can be either pure text tokens or audio tokens.

\textbf{Stage 2: Single MCTP Module Training.}
After Stage 1, the model learns both text and audio distributions. 
The objective of Stage 2 is to train the initial MCTP module to predict one subsequent token based on the output tokens and hidden states from the LLM.
This stage employs the same dataset configuration as Stage 1. We initialize the MCTP module using parameters from the final layer of the LLM, with the gradient detached from the LLM.

\textbf{Stage 3: Multiple MCTP Modules Training.}
The objective of this stage is to extend the single MCTP module to multiple MCTP modules.
Specifically, each MCTP module predicts the token at its corresponding position given the output tokens and hidden states of the previous MCTP module.
All subsequent MCTP modules are initialized using the weights of the MTCP module from Stage 2.
This stage also incorporates gradient detachment to optimize the model training process.

\textbf{Stage 4: Supervised Fine-tuning.}
After the previous three training stages, \OurMethod has acquired the ability to efficiently and accurately map text to audio.
To enable speech-to-speech dialogue capability, we then conduct supervised fine-tuning using speech QA datasets while maintaining a small amount of TTS, ASR, and text-only data to ensure training stability.
To optimize training effectiveness, different learning rates are used for the MCTP module and the main LLM.
For the speech-to-speech data, we employ an interleaved output format.
This design enforces the model to initiate audio token generation during the first forward pass, enabling synchronized decoding of the audio tokens rather than waiting until all text tokens have been generated.

\section{Limitations} 
\label{limitation}
While our approach enables efficient generation of audio tokens, the overall end-to-end latency remains above the theoretical lower bound, primarily due to the constrained generation speed of the audio decoder. Further improving the response speed of the audio decoder in end-to-end speech models is a worthwhile direction for future exploration.

\section{Data Format} 
\begin{center}
	\begin{tcolorbox}[colback=white, coltext=black, title=\textbf{Speech QA Interleaved Data Format}]
\begin{verbatim}
{
    "messages": [
    {
        "role": "user",
        "content": "<|begin_of_audio|> audio_sequence_1 <|end_of_audio|>"
    },
    {
        "role": "assistant",
        "content": "text_sequence_1 <|begin_of_audio|> 
        audio_sequence_2 <|end_of_audio|> text_sequence_2 
        <|begin_of_audio|> audio_sequence_3 <|end_of_audio|>"
    },
    {
        "role": "user",
        "content": "<|begin_of_audio|> audio_sequence_4 <|end_of_audio|>"
    },
    {
        "role": "assistant",
        "content": "text_sequence_3 <|begin_of_audio|> 
        audio_sequence_5 <|end_of_audio|> text_sequence_4 
        <|begin_of_audio|> audio_sequence_6 <|end_of_audio|>"
    }
    ]
}
\end{verbatim}
	\end{tcolorbox}
\end{center}

\begin{center}
	\begin{tcolorbox}[colback=white, coltext=black, title=\textbf{Prompt for TTS task.}]
\begin{verbatim}
{
    "messages": [
    {
        "role": "user",
        "content": "Convert the text to speech.\ntext_sequence"
    }
    ]
}
\end{verbatim}
	\end{tcolorbox}
\end{center}

\begin{center}
	\begin{tcolorbox}[colback=white, coltext=black, title=\textbf{Prompt for ASR task.}]
\begin{verbatim}
{
    "messages": [
    {
        "role": "user",
        "content": "Convert the speech to text.\n
        <|begin_of_audio|> audio_sequence <|end_of_audio|>"
    }
    ]
}
\end{verbatim}
	\end{tcolorbox}
\end{center}

\section{Figures and Tables}

\input{tables/data.tex}

\input{tables/model.tex}

\input{tables/automatic_speech_recognition.tex}

\input{tables/append_sqa.tex}

\begin{table}[h]
\caption{
We have tested VITA-Audio after aligning with ASR and TTS tasks on several text modality benchmarks. After the ASR and TTS alignment, the model’s original text understanding capabilities were indeed affected. However, this might be due to the lack of large-scale, high-quality text data used during training. Interestingly, we observed a slight improvement in performance on GSM8K, which may be due to the fact that our training dataset included a significant amount of math-related data. 
}
\begin{center}
\begin{tabular}{lcccc}
\toprule
{Model} & {MMLU\cite{hendryckstest2021}} & {GSM8K\cite{cobbe2021gsm8k}}  \\
\midrule
Qwen-7B-Instruct & 74.22 & 80.06  \\
VITA-Audio-Plus-Vanilla & 66.92 & 80.14 \\

\bottomrule

\end{tabular}
\end{center}
\end{table}

\input{tables/detail_e2e_time}

%% file: tables/data.tex
\begin{table*}[h]
\caption{
Summary of datasets used in \OurMethod for different stages.
}
\begin{center}
\begin{adjustbox}{max width=0.99\textwidth}
\begin{tabular}{c|lr|cccc}
\toprule

\multirow{2}{*}{Task} 
& \multirow{2}{*}{Name}
& \multirow{2}{*}{Total Number} & \multicolumn{4}{c}{Sampling Ratio} \\

& & & Stage 1 & Stage 2 & Stage 3 & Stage 4\\

\midrule

\multirow{12}{*}{ASR}

& WenetSpeech~\hfilll~\cite{WenetSpeech} & 10,000H & 1.0 & 1.0 & 1.0 & 0.05 \\

& Librispeech~\hfilll~\cite{Librispeech} & 1,000H & 1.0 & 1.0 & 1.0 & 0.05 \\

& Multilingual LibriSpeech~\hfilll~\cite{MultilingualLibriSpeech} & 71,506H & 1.0 & 1.0 & 1.0 & 0.05 \\

& Common Voice 17~\hfilll~\cite{Librispeech} & 2.849H & 1.0 & 1.0 & 1.0 & 0.05 \\

& MMCRSC~\hfilll~\cite{MMCRSC} & 755H & 1.0 & 1.0 & 1.0 & 0.05 \\

& GigaSpeech~\hfilll~\cite{GigaSpeech} & 10,000H & 1.0 & 1.0 & 1.0 & 0.05 \\

& People's Speech~\hfilll~\cite{PeoplesSpeech} & 1,000H & 1.0 & 1.0 & 1.0 & 0.05 \\

& VoxPopuli~\hfilll~\cite{VoxPopuli} & 543H & 1.0 & 1.0 & 1.0 & 0.05 \\

& AISHELL-1~\hfilll~\cite{AISHELL_1} & 170H& 1.0 & 1.0 & 1.0 & 0.05 \\

& AISHELL-2~\hfilll~\cite{AISHELL_2} & 1,000H & 1.0 & 1.0 & 1.0 & 0.05 \\

& AISHELL-3~\hfilll~\cite{AISHELL_3} & 85H & 1.0 & 1.0 & 1.0 & 0.05 \\

& AISHELL-4~\hfilll~\cite{AISHELL_4} & 120H & 1.0 & 1.0 & 1.0 & 0.05 \\

\midrule

\multirow{4}{*}{TTS}

& Wenetspeech4TTS~\hfilll~\cite{Wenetspeech4TTS} & 12,800H & 1.0 & 1.0 & 1.0 & 0.05 \\

& LibriTTS~\hfilll~\cite{LibriTTS} & 585H & 1.0 & 1.0 & 1.0 & 0.05 \\

& GLOBE~\hfilll~\cite{GLOBE} & 535H & 1.0 & 1.0 & 1.0 & 0.05 \\

& Emilia~\hfilll~\cite{Emilia} & 96,700H & 1.0 & 1.0 & 1.0 & 0.05 \\

\midrule

\multirow{2}{*}{Speech QA}

& VoiceAssistant-400K~\hfilll~\cite{MiniOmni} & 400K & 0.0 & 0.0 & 0.0 & 2.0 \\

& AudioQA-1.0M~\hfilll~\cite{LUCY} & 1M & 0.0 & 0.0 & 0.0 & 2.0 \\

\midrule

\multirow{9}{*}{Text QA}

& OpenHermes-2.5~\hfilll~\cite{OpenHermes-2.5} & 1M & 1.0 & 1.0 & 1.0 & 0.05 \\

& LIMA~\hfilll~\cite{LIMA} & 1K & 1.0 & 1.0 & 1.0 & 0.05 \\

& databricks-dolly-15k~\hfilll~\cite{databricks-dolly-15k} & 15K & 1.0 & 1.0 & 1.0 & 0.05 \\

& MetaMathQA~\hfilll~\cite{MetaMathQA} & 395K & 1.0 & 1.0 & 1.0 & 0.05 \\

& MathInstruct~\hfilll~\cite{MathInstruct} & 262K & 1.0 & 1.0 & 1.0 & 0.05 \\

& Orca-Math~\hfilll~\cite{Orca-Math} & 200K & 1.0 & 1.0 & 1.0 & 0.05 \\

& atlas-math-sets~\hfilll~\cite{atlas-math-sets} & 17.8M & 1.0 & 1.0 & 1.0 & 0.05 \\

& goat~\hfilll~\cite{goat} & 1.7M & 1.0 & 1.0 & 1.0 & 0.05 \\

& camel-ai-math~\hfilll~\cite{CAMEL} & 50K & 1.0 & 1.0 & 1.0 & 0.05 \\

\midrule

\multirow{8}{*}{Long Text QA}

& Long-Instruction~\hfilll~\cite{Long-Instruction-with-Paraphrasing} & 16K & 1.0 & 1.0 & 1.0 & 0.05 \\

& LongForm~\hfilll~\cite{LongForm} & 23K & 1.0 & 1.0 & 1.0 & 0.05 \\

& LongAlign-10k~\hfilll~\cite{LongAlign} & 10K & 1.0 & 1.0 & 1.0 & 0.05 \\

& LongCite-45k~\hfilll~\cite{LongCite} & 45K & 1.0 & 1.0 & 1.0 & 0.05 \\

& LongWriter-6k~\hfilll~\cite{LongWriter} & 6K & 1.0 & 1.0 & 1.0 & 0.05 \\

& LongQLoRA~\hfilll~\cite{LongQLoRA} & 39K & 1.0 & 1.0 & 1.0 & 0.05 \\

& LongAlpaca~\hfilll~\cite{LongLoRA} & 12K & 1.0 & 1.0 & 1.0 & 0.05 \\

& Long-Data-Collections~\hfilll~\cite{Long-Data-Collections} & 98K & 1.0 & 1.0 & 1.0 & 0.05 \\

\bottomrule
\end{tabular}
\end{adjustbox}
\end{center}
\label{table_data}
\end{table*}

%% file: tables/model.tex
\begin{table*}[t]
\caption{
Comparison of model structures between \OurMethod and \OurMethod-Plus.
}
\begin{center}
\begin{adjustbox}{max width=0.99\textwidth}
\begin{tabular}{l|cccc}
\toprule

{Name} & {Base LLM} & {Audio Encoder} & Audio Adapter & {Audio Decoder} \\ 

\midrule

\OurMethod & Qwen2.5-7B~\cite{Qwen2.5} & GLM-4-Voice-Tokenizer~\cite{GLM-4-Voice} & -- & GLM-4-Voice-Decoder~\cite{GLM-4-Voice} \\

\OurMethod-Plus & Qwen2.5-7B~\cite{Qwen2.5} & SenseVoiceSmall~\cite{FunAudioLLM} & MLP &GLM-4-Voice-Decoder~\cite{GLM-4-Voice} \\

\bottomrule
\end{tabular}
\end{adjustbox}
\end{center}
\label{table_model}
\end{table*}

%% file: tables/automatic_speech_recognition.tex
\begin{table*}[t]
\caption{Results on Automatic Speech Recognition~(ASR) Benchmarks.
``S$x$'' denotes the $x$-th training stage.
Compared to other methods, \textbf{\OurMethod is trained with open-source data only}.
}
\label{table_ASR_1}
\begin{center}
\begin{adjustbox}{max width=0.80\textwidth}
\begin{tabular}{c l c}
\toprule
{Datasets} & {Model} & WER (\%) $\downarrow$ \\

\midrule
\multirow{11}{*}{\begin{tabular}[c]{@{}c@{}}{LibriSpeech}~\cite{Librispeech} \\ \textit{test-clean} | \textit{test-other}\end{tabular}} 
& Qwen2-Audio-base~\hfilll~\cite{Qwen2-Audio} & \textbf{1.74} | \textbf{4.04} \\ 
& Baichuan-Audio-base~\hfilll~\cite{Baichuan-Audio} & 3.02 | 6.04 \\
& Freeze-Omni~\hfilll~\cite{Freeze-Omni} & 3.82 | 9.79 \\
& VITA-1.5 (S3)~\hfilll~\cite{VITA-1.5} & 3.40 | 7.50 \\
& LUCY (S1)~\hfilll~\cite{LUCY} & 3.36 | 8.05 \\
& Step-Audio-chat~\hfilll~\cite{Step-Audio} & 3.19 | 10.67 \\
& Qwen2.5-Omni~\hfilll~\cite{Qwen2.5-Omni} & 2.37 | 4.21 \\

\cmidrule{2-3}

& \OurMethod-Turbo & 6.29 | 12.86 \\
& \OurMethod-Vanilla & 2.98 | 8.07 \\

& \OurMethod-Plus-Boost & 3.13 | 7.07 \\
& \OurMethod-Plus-Vanilla (S1) & 1.91 | 4.29 \\
& \OurMethod-Plus-Vanilla & 2.00 | 4.60 \\

\midrule
\multirow{5}{*}{\begin{tabular}[c]{@{}c@{}}{Fleurs}~\cite{Fleurs} \\ \textit{zh} | \textit{en}\end{tabular}}
& Qwen2-Audio-base~\hfilll~\cite{Qwen2-Audio} & 3.63 | 5.20 \\ 
& Baichuan-Audio-base~\hfilll~\cite{Baichuan-Audio} & 4.15 | 8.07 \\
& Step-Audio-chat~\hfilll~\cite{Step-Audio} & 4.26 | 8.56 \\
& Qwen2.5-Omni~\hfilll~\cite{Qwen2.5-Omni} & \textbf{2.92} | \textbf{4.17} \\

\cmidrule{2-3}

& \OurMethod-Plus-Vanilla & 3.69 | 4.54 \\

\bottomrule
\end{tabular}
\end{adjustbox}
\end{center}
\end{table*}

\begin{table*}[t]
\caption{Results on Automatic Speech Recognition~(ASR) Benchmarks.
``S$x$'' denotes the $x$-th training stage.
Compared to other methods, \textbf{\OurMethod is trained with open-source data only}.
}
\label{table_ASR_2}
\begin{center}
\begin{adjustbox}{max width=0.80\textwidth}
\begin{tabular}{c l c}
\toprule
{Datasets} & {Model} & WER (\%) $\downarrow$ \\

\midrule
\multirow{11}{*}{\begin{tabular}[c]{@{}c@{}}{AISHELL-1}~\cite{AISHELL_1} \end{tabular}} 
& Qwen2-Audio-base~\hfilll~\cite{Qwen2-Audio} & 1.52 \\ 
& Baichuan-Audio-base~\hfilll~\cite{Baichuan-Audio} & 1.93 \\
& Freeze-Omni~\hfilll~\cite{Freeze-Omni} & 2.48 \\
& LUCY (S1)~\hfilll~\cite{LUCY} & 2.40 \\
& Step-Audio-chat~\hfilll~\cite{Step-Audio} & 2.14 \\
& Qwen2.5-Omni~\hfilll~\cite{Qwen2.5-Omni} & \textbf{1.13} \\

\cmidrule{2-3}
& \OurMethod-Turbo & 7.70 \\
& \OurMethod-Vanilla & 4.46 \\

& \OurMethod-Plus-Boost & 4.72 \\
& \OurMethod-Plus-Vanilla (S1) & 1.51 \\
& \OurMethod-Plus-Vanilla & 1.94 \\

\midrule
\multirow{5}{*}{\begin{tabular}[c]{@{}c@{}}{AISHELL-2} ios~\cite{AISHELL_2} \end{tabular}}
& Qwen2-Audio-base~\hfilll~\cite{Qwen2-Audio} & 3.08 \\ 
& Baichuan-Audio-base~\hfilll~\cite{Baichuan-Audio} & 3.87 \\
& Step-Audio-chat~\hfilll~\cite{Step-Audio} & 3.89 \\
& Qwen2.5-Omni~\hfilll~\cite{Qwen2.5-Omni} & \textbf{2.56} \\

\cmidrule{2-3}

& \OurMethod-Plus-Vanilla & 3.29 \\

\midrule
\multirow{11}{*}{\begin{tabular}[c]{@{}c@{}}{WenetSpeech}~\cite{WenetSpeech} \\ \textit{test-meeting} | \textit{test-net} \end{tabular}}
& Qwen2-Audio-base~\hfilll~\cite{Qwen2-Audio} & 8.40 | 7.64 \\ 
& Baichuan-Audio-base~\hfilll~\cite{Baichuan-Audio} & 13.28 | 10.13 \\
& Freeze-Omni~\hfilll~\cite{Freeze-Omni} & 13.46 | 11.80 \\
& VITA-1.5 (S3)~\hfilll~\cite{VITA-1.5} & 10.0 | 8.40 \\
& LUCY (S1)~\hfilll~\cite{LUCY} & 10.42 | 8.78 \\
& Step-Audio-chat~\hfilll~\cite{Step-Audio} & 10.83 | 9.47 \\
& Qwen2.5-Omni~\hfilll~\cite{Qwen2.5-Omni} & 7.71 | \textbf{6.04} \\

\cmidrule{2-3}

& \OurMethod-Turbo & 23.97 | 18.66 \\
& \OurMethod-Vanilla & 17.34 | 13.45 \\

& \OurMethod-Plus-Boost & 9.38 | 8.97 \\
& \OurMethod-Plus-Vanilla (S1) & \textbf{6.68} | 6.59 \\
& \OurMethod-Plus-Vanilla & 7.12 | 6.90 \\

\bottomrule
\end{tabular}
\end{adjustbox}
\end{center}
\end{table*}

%% file: tables/append_sqa.tex
\begin{table*}[t]
\caption{
More results on Spoken Question Answering~(SQA) benchmarks.
``S$x$'' denotes the $x$-th training stage of speech models.
}
\label{table_SQA_more}

\begin{center}
\begin{adjustbox}{max width=0.99\textwidth}
\begin{tabular}{lc|cc|cc|cc|cc}
\toprule

\multirow{2}{*}{Model} 
& \multirow{2}{*}{\#Params} 
& \multicolumn{2}{c|}{Llama Question}
& \multicolumn{2}{c|}{TriviaQA}
& \multicolumn{2}{c|}{Web Question} 
& \multicolumn{2}{c}{Mean} \\
& & S $\rightarrow$ T & S $\rightarrow$ S & S $\rightarrow$ T & S $\rightarrow$ S & S $\rightarrow$ T & S $\rightarrow$ S & S
$\rightarrow$ T & S $\rightarrow$ S \\

\midrule
\multicolumn{10}{c}{Proprietary Models} \\

MinMo~\hfilll~\cite{MinMo} & 7B & 78.9 & 64.1 & 48.3 & 37.5 & 55.0 & 39.9 & 60.7 & 47.2\\

\midrule
\multicolumn{10}{c}{Open-source Models} \\

Moshi~\hfilll~\cite{Moshi} & 7B & 62.3 & 21.0 & 22.8 & 7.3 & 26.6 & 9.2& 37.2 & 12.5 \\

GLM-4-Voice~\hfilll~\cite{GLM-4-Voice} & 9B & 64.7 & 50.7 & {39.1} & 26.5 & \textbf{55.0} & {39.9} & 45.3 & 31.0 \\

LUCY (S2)~\hfilll~\cite{LUCY} & 7B & 59.6 & 51.0 & 23.2 & 18.2 & 26.6 & 18.2 & 36.5 & 29.1 \\

MiniCPM-o2.6~\hfilll~\cite{yao2024minicpm}& 7B & - & 61.0 & - & 40.0 & - & 40.2 & - & 47.0 \\

Llama-Omni~\hfilll~\cite{LLaMA-Omni} & 7B & - & 45.3 & - & 22.9 & - & 10.7 & - & 26.3 \\

\midrule
\OurMethod-Boost & 7B & 68.7 & 60.3 & 30.5 & 29.3 & 32.9 & 30.4 & 44.0 & 40.0 \\

\OurMethod-Vanilla & 7B & 71.3 & 66.3 & 31.9 & 30.1 & 33.5 & 31.4 & 45.6 & 42.6 \\

\OurMethod-Plus-Boost & 7B & \textbf{76.3} & 64.6 & 43.6 & 39.5 & 44.2 & 40.0 & 54.7 & 48.0 \\

\OurMethod-Plus-Vanilla & 7B & 75.6 & \textbf{68.0} & \textbf{45.9} & \textbf{42.7} & 45.0 & \textbf{41.7} & \textbf{55.5} & \textbf{50.8} \\

\bottomrule
\end{tabular}
\end{adjustbox}
\end{center}

\end{table*}

%% file: tables/detail_e2e_time.tex
\begin{table}[h]
\caption{
Generation time (ms) of the first audio segment under different inference modes in streaming inference.
To enable more real-time speech generation, we progressively increase the number of steps in the flow matching model during streaming inference.
The table shows the decoding time when the sampling step of the flow matching model is set to $1$.
}
\begin{center}
\begin{tabular}{lcccc}
\toprule
{Inference Mode} & {Audio Encoder} & {First Audio Token Chunk} & {Audio Decoder} & {Sum} \\
\midrule
VITA-Audio-Boost & 39 & 53 & 151 & 243 \\
VITA-Audio-Vanilla & 39 & 236 & 151 & 426 \\

\bottomrule
\label{e2e_time}
\end{tabular}
\end{center}
\end{table}

%% file: neurips_2025.bbl
\begin{thebibliography}{10}

\bibitem{FunAudioLLM}
Keyu An, Qian Chen, Chong Deng, Zhihao Du, Changfeng Gao, Zhifu Gao, Yue Gu, Ting He, Hangrui Hu, Kai Hu, Shengpeng Ji, Yabin Li, Zerui Li, Heng Lu, Haoneng Luo, Xiang Lv, Bin Ma, Ziyang Ma, Chongjia Ni, Changhe Song, Jiaqi Shi, Xian Shi, Hao Wang, Wen Wang, Yuxuan Wang, Zhangyu Xiao, Zhijie Yan, Yexin Yang, Bin Zhang, Qinglin Zhang, Shiliang Zhang, Nan Zhao, and Siqi Zheng.
\newblock {FunAudioLLM: Voice Understanding and Generation Foundation Models for Natural Interaction Between Humans and LLMs}.
\newblock {\em arXiv:2407.04051}, 2024.

\bibitem{Seed-TTS}
Philip Anastassiou, Jiawei Chen, Jitong Chen, Yuanzhe Chen, Zhuo Chen, Ziyi Chen, Jian Cong, Lelai Deng, Chuang Ding, Lu~Gao, Mingqing Gong, Peisong Huang, Qingqing Huang, Zhiying Huang, Yuanyuan Huo, Dongya Jia, Chumin Li, Feiya Li, Hui Li, Jiaxin Li, Xiaoyang Li, Xingxing Li, Lin Liu, Shouda Liu, Sichao Liu, Xudong Liu, Yuchen Liu, Zhengxi Liu, Lu~Lu, Junjie Pan, Xin Wang, Yuping Wang, Yuxuan Wang, Zhen Wei, Jian Wu, Chao Yao, Yifeng Yang, Yuanhao Yi, Junteng Zhang, Qidi Zhang, Shuo Zhang, Wenjie Zhang, Yang Zhang, Zilin Zhao, Dejian Zhong, and Xiaobin Zhuang.
\newblock {Seed-TTS: A Family of High-Quality Versatile Speech Generation Models}.
\newblock {\em arXiv:2406.02430}, 2024.

\bibitem{LongAlign}
Yushi Bai, Xin Lv, Jiajie Zhang, Yuze He, Ji~Qi, Lei Hou, Jie Tang, Yuxiao Dong, and Juanzi Li.
\newblock {LongAlign: A Recipe for Long Context Alignment of Large Language Models}.
\newblock 2024.

\bibitem{LongWriter}
Yushi Bai, Jiajie Zhang, Xin Lv, Linzhi Zheng, Siqi Zhu, Lei Hou, Yuxiao Dong, Jie Tang, and Juanzi Li.
\newblock {LongWriter: Unleashing 10,000+ Word Generation from Long Context LLMs}.
\newblock 2024.

\bibitem{Web-Questions}
Jonathan Berant, Andrew Chou, Roy Frostig, and Percy Liang.
\newblock {Semantic Parsing on Freebase from Question-Answer Pairs}.
\newblock In {\em Conference on Empirical Methods in Natural Language Processing (EMNLP)}, 2013.

\bibitem{AISHELL_1}
Hui Bu, Jiayu Du, Xingyu Na, Bengu Wu, and Hao Zheng.
\newblock {AISHELL-1: An Open-Source Mandarin Speech Corpus and a Speech Recognition Baseline}.
\newblock {\em Conference of the Oriental Chapter of the International Coordinating Committee on Speech Databases and Speech I/O Systems and Assessment (O-COCOSDA)}, 2018.

\bibitem{LAS}
William Chan, Navdeep Jaitly, Quoc~V Le, and Vinyals {Google Brain}.
\newblock {Listen, Attend and Spell}.
\newblock 2015.

\bibitem{GigaSpeech}
Guoguo Chen, Shuzhou Chai, Guanbo Wang, Jiayu Du, Wei~Qiang Zhang, Chao Weng, Dan Su, Daniel Povey, Jan Trmal, Junbo Zhang, Mingjie Jin, Sanjeev Khudanpur, Shinji Watanabe, Shuaijiang Zhao, Wei Zou, Xiangang Li, Xuchen Yao, Yongqing Wang, Zhao You, and Zhiyong Yan.
\newblock {GigaSpeech: An Evolving, Multi-Domain ASR Corpus with 10,000 Hours of Transcribed Audio}.
\newblock In {\em Proceedings of the Annual Conference of the International Speech Communication Association (Interspeech)}, 2021.

\bibitem{MinMo}
Qian Chen, Yafeng Chen, Yanni Chen, Mengzhe Chen, Yingda Chen, Chong Deng, Zhihao Du, Ruize Gao, Changfeng Gao, Zhifu Gao, Yabin Li, Xiang Lv, Jiaqing Liu, Haoneng Luo, Bin Ma, Chongjia Ni, Xian Shi, Jialong Tang, Hui Wang, Hao Wang, Wen Wang, Yuxuan Wang, Yunlan Xu, Fan Yu, Zhijie Yan, Yexin Yang, Baosong Yang, Xian Yang, Guanrou Yang, Tianyu Zhao, Qinglin Zhang, Shiliang Zhang, Nan Zhao, Pei Zhang, Chong Zhang, and Jinren Zhou.
\newblock {MinMo: A Multimodal Large Language Model for Seamless Voice Interaction}.
\newblock {\em arXiv:2501.06282}, 2025.

\bibitem{SLAM-Omni}
Wenxi Chen, Ziyang Ma, Ruiqi Yan, Yuzhe Liang, Xiquan Li, Ruiyang Xu, Zhikang Niu, Yanqiao Zhu, Yifan Yang, Zhanxun Liu, Kai Yu, Yuxuan Hu, Jinyu Li, Yan Lu, Shujie Liu, and Xie Chen.
\newblock {SLAM-Omni: Timbre-Controllable Voice Interaction System with Single-Stage Training}.
\newblock {\em arXiv:2412.15649}, 2024.

\bibitem{LongLoRA}
Yukang Chen, Shengju Qian, Haotian Tang, Xin Lai, Zhijian Liu, Song Han, and Jiaya Jia.
\newblock {LongLoRA: Efficient Fine-Tuning of Long-Context Large Language Models}.
\newblock 2023.

\bibitem{Qwen2-Audio}
Yunfei Chu, Jin Xu, Qian Yang, Haojie Wei, Xipin Wei, Zhifang Guo, Yichong Leng, Yuanjun Lv, Jinzheng He, Junyang Lin, Chang Zhou, and Jingren Zhou.
\newblock {Qwen2-Audio Technical Report}.
\newblock {\em arXiv:2407.10759}, 2024.

\bibitem{cobbe2021gsm8k}
Karl Cobbe, Vineet Kosaraju, Mohammad Bavarian, Mark Chen, Heewoo Jun, Lukasz Kaiser, Matthias Plappert, Jerry Tworek, Jacob Hilton, Reiichiro Nakano, Christopher Hesse, and John Schulman.
\newblock Training verifiers to solve math word problems.
\newblock {\em arXiv preprint arXiv:2110.14168}, 2021.

\bibitem{Long-Data-Collections}
Together Computer.
\newblock {Long Data Collections}, 2023.

\bibitem{Fleurs}
Alexis Conneau, Min Ma, Simran Khanuja, Yu~Zhang, Vera Axelrod, Siddharth Dalmia, Jason Riesa, Clara Rivera, and Ankur Bapna.
\newblock {FLEURS: FEW-Shot Learning Evaluation of Universal Representations of Speech}.
\newblock {\em 2022 IEEE Spoken Language Technology Workshop, SLT 2022 - Proceedings}, 2023.

\bibitem{databricks-dolly-15k}
Mike Conover, Matt Hayes, Ankit Mathur, Jianwei Xie, Jun Wan, Sam Shah, Ali Ghodsi, Patrick Wendell, Matei Zaharia, and Reynold Xin.
\newblock {Free Dolly: Introducing the World's First Truly Open Instruction-Tuned LLM}, 2023.

\bibitem{FlashAttention-2}
Tri Dao.
\newblock {FlashAttention-2: Faster Attention with Better Parallelism and Work Partitioning}.
\newblock {\em The International Conference on Learning Representations (ICLR)}, 2023.

\bibitem{DeepSeek-V3}
DeepSeek-AI, Aixin Liu, Bei Feng, Bing Xue, Bingxuan Wang, Bochao Wu, Chengda Lu, Chenggang Zhao, Chengqi Deng, Chenyu Zhang, Chong Ruan, Damai Dai, Daya Guo, Dejian Yang, Deli Chen, Dongjie Ji, Erhang Li, Fangyun Lin, Fucong Dai, Fuli Luo, Guangbo Hao, Guanting Chen, Guowei Li, H.~Zhang, Han Bao, Hanwei Xu, Haocheng Wang, Haowei Zhang, Honghui Ding, Huajian Xin, Huazuo Gao, Hui Li, Hui Qu, J.~L. Cai, Jian Liang, Jianzhong Guo, Jiaqi Ni, Jiashi Li, Jiawei Wang, Jin Chen, Jingchang Chen, Jingyang Yuan, Junjie Qiu, Junlong Li, Junxiao Song, Kai Dong, Kai Hu, Kaige Gao, Kang Guan, Kexin Huang, Kuai Yu, Lean Wang, Lecong Zhang, Lei Xu, Leyi Xia, Liang Zhao, Litong Wang, Liyue Zhang, Meng Li, Miaojun Wang, Mingchuan Zhang, Minghua Zhang, Minghui Tang, Mingming Li, Ning Tian, Panpan Huang, Peiyi Wang, Peng Zhang, Qiancheng Wang, Qihao Zhu, Qinyu Chen, Qiushi Du, R.~J. Chen, R.~L. Jin, Ruiqi Ge, Ruisong Zhang, Ruizhe Pan, Runji Wang, Runxin Xu, Ruoyu Zhang, Ruyi Chen, S.~S. Li, Shanghao Lu, Shangyan Zhou, Shanhuang
  Chen, Shaoqing Wu, Shengfeng Ye, Shengfeng Ye, Shirong Ma, Shiyu Wang, Shuang Zhou, Shuiping Yu, Shunfeng Zhou, Shuting Pan, T.~Wang, Tao Yun, Tian Pei, Tianyu Sun, W.~L. Xiao, Wangding Zeng, Wanjia Zhao, Wei An, Wen Liu, Wenfeng Liang, Wenjun Gao, Wenqin Yu, Wentao Zhang, X.~Q. Li, Xiangyue Jin, Xianzu Wang, Xiao Bi, Xiaodong Liu, Xiaohan Wang, Xiaojin Shen, Xiaokang Chen, Xiaokang Zhang, Xiaosha Chen, Xiaotao Nie, Xiaowen Sun, Xiaoxiang Wang, Xin Cheng, Xin Liu, Xin Xie, Xingchao Liu, Xingkai Yu, Xinnan Song, Xinxia Shan, Xinyi Zhou, Xinyu Yang, Xinyuan Li, Xuecheng Su, Xuheng Lin, Y.~K. Li, Y.~Q. Wang, Y.~X. Wei, Y.~X. Zhu, Yang Zhang, Yanhong Xu, Yanhong Xu, Yanping Huang, Yao Li, Yao Zhao, Yaofeng Sun, Yaohui Li, Yaohui Wang, Yi~Yu, Yi~Zheng, Yichao Zhang, Yifan Shi, Yiliang Xiong, Ying He, Ying Tang, Yishi Piao, Yisong Wang, Yixuan Tan, Yiyang Ma, Yiyuan Liu, Yongqiang Guo, Yu~Wu, Yuan Ou, Yuchen Zhu, Yuduan Wang, Yue Gong, Yuheng Zou, Yujia He, Yukun Zha, Yunfan Xiong, Yunxian Ma, Yuting Yan, Yuxiang
  Luo, Yuxiang You, Yuxuan Liu, Yuyang Zhou, Z.~F. Wu, Z.~Z. Ren, Zehui Ren, Zhangli Sha, Zhe Fu, Zhean Xu, Zhen Huang, Zhen Zhang, Zhenda Xie, Zhengyan Zhang, Zhewen Hao, Zhibin Gou, Zhicheng Ma, Zhigang Yan, Zhihong Shao, Zhipeng Xu, Zhiyu Wu, Zhongyu Zhang, Zhuoshu Li, Zihui Gu, Zijia Zhu, Zijun Liu, Zilin Li, Ziwei Xie, Ziyang Song, Ziyi Gao, and Zizheng Pan.
\newblock {DeepSeek-V3 Technical Report}.
\newblock 2024.

\bibitem{Moshi}
Alexandre D{\'{e}}fossez, Laurent Mazar{\'{e}}, Manu Orsini, Am{\'{e}}lie Royer, Patrick P{\'{e}}rez, Herv{\'{e}} J{\'{e}}gou, Edouard Grave, and Neil Zeghidour.
\newblock {Moshi: A Speech-Text Foundation Model for Real-Time Dialogue}.
\newblock {\em arXiv:2410.00037}, 2024.

\bibitem{AISHELL_2}
Jiayu Du, Xingyu Na, Xuechen Liu, and Hui Bu.
\newblock {AISHELL-2: Transforming Mandarin ASR Research into Industrial Scale}.
\newblock {\em arXiv:1808.10583}, 2018.

\bibitem{CosyVoice}
Zhihao Du, Qian Chen, Shiliang Zhang, Kai Hu, Heng Lu, Yexin Yang, Hangrui Hu, Siqi Zheng, Yue Gu, Ziyang Ma, Zhifu Gao, and Zhijie Yan.
\newblock {CosyVoice: A Scalable Multilingual Zero-Shot Text-To-Speech Synthesizer Based on Supervised Semantic Tokens}.
\newblock {\em arXiv:2407.05407}, 2024.

\bibitem{CosyVoice2}
Zhihao Du, Yuxuan Wang, Qian Chen, Xian Shi, Xiang Lv, Tianyu Zhao, Zhifu Gao, Yexin Yang, Changfeng Gao, Hui Wang, Fan Yu, Huadai Liu, Zhengyan Sheng, Yue Gu, Chong Deng, Wen Wang, Shiliang Zhang, Zhijie Yan, and Jingren Zhou.
\newblock {CosyVoice 2: Scalable Streaming Speech Synthesis with Large Language Models}.
\newblock {\em arXiv:2412.10117}, 2024.

\bibitem{LLaMA-Omni}
Qingkai Fang, Shoutao Guo, Yan Zhou, Zhengrui Ma, Shaolei Zhang, and Yang Feng.
\newblock {LLaMA-Omni: Seamless Speech Interaction with Large Language Models}.
\newblock {\em arXiv:2409.06666}, 2024.

\bibitem{VITA}
Chaoyou Fu, Haojia Lin, Zuwei Long, Yunhang Shen, Meng Zhao, Yifan Zhang, Xiong Wang, Di~Yin, Long Ma, Xiawu Zheng, Ran He, Rongrong Ji, Yunsheng Wu, Caifeng Shan, and Xing Sun.
\newblock {VITA: Towards Open-Source Interactive Omni Multimodal LLM}.
\newblock {\em arXiv:2408.05211}, 2024.

\bibitem{VITA-1.5}
Chaoyou Fu, Haojia Lin, Xiong Wang, Yi-Fan Zhang, Yunhang Shen, Xiaoyu Liu, Yangze Li, Zuwei Long, Heting Gao, Ke~Li, Xiawu Zheng, Rongrong Ji, Xing Sun, Caifeng Shan, and Ran He.
\newblock {VITA-1.5: Towards GPT-4o Level Real-Time Vision and Speech Interaction}.
\newblock {\em arXiv:2501.01957}, 2025.

\bibitem{AISHELL_4}
Yihui Fu, Luyao Cheng, Shubo Lv, Yukai Jv, Yuxiang Kong, Zhuo Chen, Yanxin Hu, Lei Xie, Jian Wu, Hui Bu, Xin Xu, Jun Du, and Jingdong Chen.
\newblock {AISHELL-4: An Open Source Dataset for Speech Enhancement, Separation, Recognition and Speaker Diarization in Conference Scenario}.
\newblock In {\em Proceedings of the Annual Conference of the International Speech Communication Association (Interspeech)}, 2021.

\bibitem{PeoplesSpeech}
Daniel Galvez, Greg Diamos, Juan Ciro, Juan~Felipe Cer{\'{o}}n, Keith Achorn, Anjali Gopi, David Kanter, Maximilian Lam, Mark Mazumder, and Vijay~Janapa Reddi.
\newblock {the People's Speech: A Large-Scale Diverse English Speech Recognition Dataset for Commercial Usage}.
\newblock 2021.

\bibitem{LUCY}
Heting Gao, Hang Shao, Xiong Wang, Chaofan Qiu, Yunhang Shen, Siqi Cai, Yuchen Shi, Zihan Xu, Zuwei Long, Yike Zhang, Shaoqi Dong, Chaoyou Fu, Ke~Li, Long Ma, and Xing Sun.
\newblock {LUCY: Linguistic Understanding and Control Yielding Early Stage of Her}.
\newblock {\em arXiv:2501.16327}, 2025.

\bibitem{Paraformer}
Zhifu Gao, Shiliang Zhang, Ian McLoughlin, and Zhijie Yan.
\newblock {Paraformer: Fast and Accurate Parallel Transformer for Non-Autoregressive End-To-End Speech Recognition}.
\newblock In {\em Proceedings of the Annual Conference of the International Speech Communication Association (Interspeech)}, 2022.

\bibitem{Emilia}
Haorui He, Zengqiang Shang, Chaoren Wang, Xuyuan Li, Yicheng Gu, Hua Hua, Liwei Liu, Chen Yang, Jiaqi Li, Peiyang Shi, Yuancheng Wang, Kai Chen, Pengyuan Zhang, and Zhizheng Wu.
\newblock {Emilia: An Extensive, Multilingual, and Diverse Speech Dataset for Large-Scale Speech Generation}.
\newblock {\em arXiv:2501.15907}, 2024.

\bibitem{hendryckstest2021}
Dan Hendrycks, Collin Burns, Steven Basart, Andy Zou, Mantas Mazeika, Dawn Song, and Jacob Steinhardt.
\newblock Measuring massive multitask language understanding.
\newblock {\em Proceedings of the International Conference on Learning Representations (ICLR)}, 2021.

\bibitem{AudioGPT}
Rongjie Huang, Mingze Li, Dongchao Yang, Jiatong Shi, Xuankai Chang, Zhenhui Ye, Yuning Wu, Zhiqing Hong, Jiawei Huang, Jinglin Liu, Yi~Ren, Yuexian Zou, Zhou Zhao, and Shinji Watanabe.
\newblock {AudioGPT: Understanding and Generating Speech, Music, Sound, and Talking Head}.
\newblock In {\em AAAI Conference on Artificial Intelligence (AAAI)}, 2023.

\bibitem{TriviaQA}
Mandar Joshi, Eunsol Choi, Daniel~S. Weld, and Luke Zettlemoyer.
\newblock {TriviaQA: A Large Scale Distantly Supervised Challenge Dataset for Reading Comprehension}.
\newblock {\em Proceedings of the Annual Meeting of the Association for Computational Linguistics (ACL)}, 2017.

\bibitem{Joint-CTC-attention}
Suyoun Kim, Takaaki Hori, and Shinji Watanabe.
\newblock {Joint CTC-Attention Based End-To-End Speech Recognition Using Multi-Task Learning}.
\newblock In {\em International Conference on Acoustics, Speech and Signal Processing (ICASSP)}, 2017.

\bibitem{LongForm}
Abdullatif K{\"{o}}ksal, Timo Schick, Anna Korhonen, and Hinrich Sch{\"{u}}tze.
\newblock {LongForm: Optimizing Instruction Tuning for Long Text Generation with Corpus Extraction}.
\newblock 2023.

\bibitem{CAMEL}
Guohao Li, Hasan Abed Al~Kader Hammoud, Hani Itani, Dmitrii Khizbullin, and Bernard Ghanem.
\newblock {CAMEL: Communicative Agents for "Mind" Exploration of Large Language Model Society}.
\newblock 2023.

\bibitem{Baichuan-Audio}
Tianpeng Li, Jun Liu, Tao Zhang, Yuanbo Fang, Da~Pan, Mingrui Wang, Zheng Liang, Zehuan Li, Mingan Lin, Guosheng Dong, Jianhua Xu, Haoze Sun, Zenan Zhou, and Weipeng Chen.
\newblock {Baichuan-Audio: A Unified Framework for End-To-End Speech Interaction}.
\newblock {\em arXiv:2502.17239}, 2025.

\bibitem{EAGLE}
Yuhui Li, Fangyun Wei, Chao Zhang, and Hongyang Zhang.
\newblock {EAGLE: Speculative Sampling Requires Rethinking Feature Uncertainty}.
\newblock {\em International Conference on Machine Learning (ICML)}, 2024.

\bibitem{Wenetspeech4TTS}
Linhan Ma, Dake Guo, Kun Song, Yuepeng Jiang, Shuai Wang, Liumeng Xue, Weiming Xu, Huan Zhao, Binbin Zhang, and Lei Xie.
\newblock {WenetSpeech4TTS: A 12,800-Hour Mandarin TTS Corpus for Large Speech Generation Model Benchmark}.
\newblock In {\em Proceedings of the Annual Conference of the International Speech Communication Association (Interspeech)}, 2024.

\bibitem{MMCRSC}
Ltd. {Magic Data Technology Co.}
\newblock {MAGICDATa Mandarin Chinese Read Speech Corpus}, 2019.

\bibitem{Orca-Math}
Arindam Mitra, Hamed Khanpour, Corby Rosset, and Ahmed Awadallah.
\newblock {Orca-Math: Unlocking the Potential of SLMs in Grade School Math}.
\newblock 2024.

\bibitem{Llama-Question}
Eliya Nachmani, Alon Levkovitch, Roy Hirsch, Julian Salazar, Chulayuth Asawaroengchai, Soroosh Mariooryad, Ehud Rivlin, R.~J. Skerry-Ryan, and Michelle~Tadmor Ramanovich.
\newblock {Spoken Question Answering and Speech Continuation Using Spectrogram-Powered Llm}.
\newblock In {\em The International Conference on Learning Representations (ICLR)}, 2024.

\bibitem{Librispeech}
Vassil Panayotov, Guoguo Chen, Daniel Povey, and Sanjeev Khudanpur.
\newblock {Librispeech: An ASR Corpus Based on Public Domain Audio Books}.
\newblock In {\em International Conference on Acoustics, Speech and Signal Processing (ICASSP)}, 2015.

\bibitem{MultilingualLibriSpeech}
Vineel Pratap, Qiantong Xu, Anuroop Sriram, Gabriel Synnaeve, and Ronan Collobert.
\newblock {MLS: A Large-Scale Multilingual Dataset for Speech Research}.
\newblock In {\em Proceedings of the Annual Conference of the International Speech Communication Association (Interspeech)}, 2020.

\bibitem{Qwen2.5}
Qwen, :, An~Yang, Baosong Yang, Beichen Zhang, Binyuan Hui, Bo~Zheng, Bowen Yu, Chengyuan Li, Dayiheng Liu, Fei Huang, Haoran Wei, Huan Lin, Jian Yang, Jianhong Tu, Jianwei Zhang, Jianxin Yang, Jiaxi Yang, Jingren Zhou, Junyang Lin, Kai Dang, Keming Lu, Keqin Bao, Kexin Yang, Le~Yu, Mei Li, Mingfeng Xue, Pei Zhang, Qin Zhu, Rui Men, Runji Lin, Tianhao Li, Tianyi Tang, Tingyu Xia, Xingzhang Ren, Xuancheng Ren, Yang Fan, Yang Su, Yichang Zhang, Yu~Wan, Yuqiong Liu, Zeyu Cui, Zhenru Zhang, and Zihan Qiu.
\newblock {Qwen2.5 Technical Report}.
\newblock 2024.

\bibitem{Whisper}
Alec Radford, Jong~Wook Kim, Tao Xu, Greg Brockman, Christine McLeavey, and Ilya Sutskever.
\newblock {Robust Speech Recognition Via Large-Scale Weak Supervision}.
\newblock {\em International Conference on Machine Learning (PMLR)}, 2022.

\bibitem{Long-VITA}
Yunhang Shen, Chaoyou Fu, Shaoqi Dong, Xiong Wang, Peixian Chen, Mengdan Zhang, Haoyu Cao, Ke~Li, Xiawu Zheng, Yan Zhang, Yiyi Zhou, Rongrong Ji, and Xing Sun.
\newblock {Long-VITA: Scaling Large Multi-Modal Models to 1 Million Tokens with Leading Short-Context Accuracy}.
\newblock {\em arXiv:2502.05177}, 2025.

\bibitem{AISHELL_3}
Yao Shi, Hui Bu, Xin Xu, Shaoji Zhang, and Ming Li.
\newblock {AISHELL-3: A Multi-Speaker Mandarin TTS Corpus and the Baselines}.
\newblock In {\em Proceedings of the Annual Conference of the International Speech Communication Association (Interspeech)}, 2020.

\bibitem{Step-Audio}
Step-audio Team.
\newblock {Step-Audio: Unified Understanding and Generation in Intelligent Speech Interaction}.
\newblock {\em arXiv:2502.11946}, 2024.

\bibitem{OpenHermes-2.5}
Teknium.
\newblock {OpenHermes 2.5: An Open Dataset of Synthetic Data for Generalist LLM Assistants}, 2023.

\bibitem{goat}
Liu Tiedong.
\newblock Goat, 2023.

\bibitem{atlas-math-sets}
Atlas Unified.
\newblock Atlas math sets, 2023.

\bibitem{VoxPopuli}
Changhan Wang, Morgane Rivi{\`{e}}re, Ann Lee, Anne Wu, Chaitanya Talnikar, Daniel Haziza, Mary Williamson, Juan Pino, and Emmanuel Dupoux.
\newblock {VoxPopuli: A Large-Scale Multilingual Speech Corpus for Representation Learning, Semi-Supervised Learning and Interpretation}.
\newblock In {\em International Joint Conference on Natural Language Processing (IJCNLP)}, 2021.

\bibitem{MThreads-full-duplex}
Peng Wang, Songshuo Lu, Yaohua Tang, Sijie Yan, Yuanjun Xiong, Wei Xia, and Mthreads Ai.
\newblock {a Full-Duplex Speech Dialogue Scheme Based on Large Language Models}.

\bibitem{GLOBE}
Wenbin Wang, Yang Song, and Sanjay Jha.
\newblock {GLOBE: A High-Quality English Corpus with Global Accents for Zero-Shot Speaker Adaptive Text-To-Speech}.
\newblock In {\em Proceedings of the Annual Conference of the International Speech Communication Association (Interspeech)}, 2024.

\bibitem{Freeze-Omni}
Xiong Wang, Yangze Li, Chaoyou Fu, Lei Xie, Ke~Li, Xing Sun, and Long Ma.
\newblock {Freeze-Omni: A Smart and Low Latency Speech-To-Speech Dialogue Model with Frozen LLM}.
\newblock {\em arXiv:2411.00774}, 2024.

\bibitem{Transformers}
Thomas Wolf, Lysandre Debut, Victor Sanh, Julien Chaumond, Clement Delangue, Anthony Moi, Pierric Cistac, Tim Rault, R{\'{e}}mi Louf, Morgan Funtowicz, Joe Davison, Sam Shleifer, Patrick von Platen, Clara Ma, Yacine Jernite, Julien Plu, Canwen Xu, Teven~Le Scao, Sylvain Gugger, Mariama Drame, Quentin Lhoest, and Alexander~M. Rush.
\newblock {HuggingFace's Transformers: State-Of-The-Art Natural Language Processing}.
\newblock 2019.

\bibitem{MiniOmni}
Zhifei Xie and Changqiao Wu.
\newblock {Mini-Omni: Language Models Can Hear, Talk While Thinking in Streaming}.
\newblock {\em arXiv:2408.16725}, 2024.

\bibitem{Qwen2.5-Omni}
Jin Xu, Zhifang Guo, Jinzheng He, Hangrui Hu, Ting He, Shuai Bai, Keqin Chen, Jialin Wang, Yang Fan, Kai Dang, Bin Zhang, Xiong Wang, Yunfei Chu, and Junyang Lin.
\newblock {Qwen2.5-Omni Technical Report}.
\newblock {\em arXiv:2503.20215}, 2025.

\bibitem{LongQLoRA}
Jianxin Yang.
\newblock {LongQLoRA: Efficient and Effective Method to Extend Context Length of Large Language Models}.
\newblock 2023.

\bibitem{yao2024minicpm}
Yuan Yao, Tianyu Yu, Ao~Zhang, Chongyi Wang, Junbo Cui, Hongji Zhu, Tianchi Cai, Haoyu Li, Weilin Zhao, Zhihui He, et~al.
\newblock Minicpm-v: A gpt-4v level mllm on your phone.
\newblock {\em arXiv preprint arXiv:2408.01800}, 2024.

\bibitem{MetaMathQA}
Longhui Yu, Weisen Jiang, Han Shi, Jincheng Yu, Zhengying Liu, Yu~Zhang, James~T. Kwok, Zhenguo Li, Adrian Weller, and Weiyang Liu.
\newblock {Metamath: Bootstrap Your Own Mathematical Questions for Large Language Models}.
\newblock {\em The International Conference on Learning Representations (ICLR)}, 2024.

\bibitem{Long-Instruction-with-Paraphrasing}
Yijiong Yu.
\newblock {"Paraphrasing the Original Text" Makes High Accuracy Long-Context QA}.
\newblock 2023.

\bibitem{MathInstruct}
Xiang Yue, Xingwei Qu, Ge~Zhang, Yao Fu, Wenhao Huang, Huan Sun, Yu~Su, and Wenhu Chen.
\newblock {Mammoth: Building Math Generalist Models Through Hybrid Instruction Tuning}.
\newblock {\em The International Conference on Learning Representations (ICLR)}, 2024.

\bibitem{LibriTTS}
Heiga Zen, Viet Dang, Rob Clark, Yu~Zhang, Ron~J. Weiss, Ye~Jia, Zhifeng Chen, and Yonghui Wu.
\newblock {Libritts: A Corpus Derived from LibriSpeech for Text-To-Speech}.
\newblock In {\em Proceedings of the Annual Conference of the International Speech Communication Association (Interspeech)}, 2019.

\bibitem{GLM-4-Voice}
Aohan Zeng, Zhengxiao Du, Mingdao Liu, Kedong Wang, Shengmin Jiang, Lei Zhao, Yuxiao Dong, and Jie Tang.
\newblock {GLM-4-Voice: Towards Intelligent and Human-like End-To-End Spoken Chatbot}.
\newblock {\em arXiv:2412.02612}, 2024.

\bibitem{WenetSpeech}
Binbin Zhang, Hang Lv, Pengcheng Guo, Qijie Shao, Chao Yang, Lei Xie, Xin Xu, Hui Bu, Xiaoyu Chen, Chenchen Zeng, Di~Wu, and Zhendong Peng.
\newblock {WenetSpeech: A 10000+ Hours Multi-Domain Mandarin Corpus for Speech Recognition}.
\newblock In {\em International Conference on Acoustics, Speech and Signal Processing (ICASSP)}, 2022.

\bibitem{LongCite}
Jiajie Zhang, Yushi Bai, Xin Lv, Wanjun Gu, Danqing Liu, Minhao Zou, Shulin Cao, Lei Hou, Yuxiao Dong, Ling Feng, and Juanzi Li.
\newblock {LongCite: Enabling LLMs to Generate Fine-Grained Citations in Long-Context QA}.
\newblock 2024.

\bibitem{InternLM-XComposer2.5-OmniLive}
Pan Zhang, Xiaoyi Dong, Yuhang Cao, Yuhang Zang, Rui Qian, Xilin Wei, Lin Chen, Yifei Li, Junbo Niu, Shuangrui Ding, Qipeng Guo, Haodong Duan, Xin Chen, Han Lv, Zheng Nie, Min Zhang, Bin Wang, Wenwei Zhang, Xinyue Zhang, Jiaye Ge, Wei Li, Jingwen Li, Zhongying Tu, Conghui He, Xingcheng Zhang, Kai Chen, Yu~Qiao, Dahua Lin, and Jiaqi Wang.
\newblock {InternLM-XComposer2.5-OmniLive: A Comprehensive Multimodal System for Long-Term Streaming Video and Audio Interactions}.
\newblock 2024.

\bibitem{LIMA}
Chunting Zhou, Pengfei Liu, Puxin Xu, Srini Iyer, Jiao Sun, Yuning Mao, Xuezhe Ma, Avia Efrat, Ping Yu, Lili Yu, Susan Zhang, Gargi Ghosh, Mike Lewis, Luke Zettlemoyer, and Omer Levy.
\newblock {LIMA: Less Is More for Alignment}.
\newblock 2023.

\end{thebibliography}
